\documentclass[twocolumn, switch]{article} 

\usepackage{preprint}

\usepackage{amsmath, amsthm, amssymb, amsfonts}
\usepackage{bm}
\usepackage{mathtools}

\usepackage[numbers,square]{natbib}
\usepackage[utf8]{inputenc}	
\usepackage[T1]{fontenc}	
\usepackage{xcolor}		
\usepackage[colorlinks = true,
            linkcolor = blue,
            urlcolor  = blue,
            citecolor = blue,
            anchorcolor = black]{hyperref}	
\usepackage{booktabs} 		
\usepackage{nicefrac}		
\usepackage{microtype}		
\usepackage{lineno}		
\usepackage{float}			

\usepackage{newfloat}
\DeclareFloatingEnvironment[name={Supplementary Figure}]{suppfigure}
\usepackage{sidecap}
\sidecaptionvpos{figure}{c}

\usepackage{titlesec}
\titlespacing\section{0pt}{12pt plus 3pt minus 3pt}{1pt plus 1pt minus 1pt}
\titlespacing\subsection{0pt}{10pt plus 3pt minus 3pt}{1pt plus 1pt minus 1pt}
\titlespacing\subsubsection{0pt}{8pt plus 3pt minus 3pt}{1pt plus 1pt minus 1pt}

\usepackage{graphics}
\usepackage{graphicx}
\usepackage{epstopdf}
\usepackage{epsfig}
\usepackage{hyperref}
\usepackage{color}
\usepackage{multirow}
\usepackage{hhline}
\usepackage[flushleft]{threeparttable}
\usepackage{subfigure}
\usepackage{amsthm,amsmath,amssymb}
\usepackage{mathrsfs}
\usepackage{color}
\usepackage{comment}

\title{Towards Mesh Saliency Detection in 6 Degrees of Freedom}

\usepackage{authblk}

\author{Xiaoying Ding}
\author{Zhenzhong Chen\thanks{\tt{zzchen@ieee.org}}}
\affil{School of Remote Sensing and Information Engineering, Wuhan University}

\begin{document}

\twocolumn[ 
  \begin{@twocolumnfalse} 
  
\maketitle

\begin{abstract}
Traditional 3D mesh saliency detection algorithms and corresponding databases were proposed under several constraints such as providing limited viewing directions and not taking the subject's movement into consideration. In this work, a novel 6DoF mesh saliency database is developed which provides both the subject's 6DoF data and eye-movement data. Different from traditional databases, subjects in the experiment are allowed to move freely to observe 3D meshes in a virtual reality environment. Based on the database, we first analyze the inter-observer variation and the influence of viewing direction towards subject's visual attention, then we provide further investigations about the subject's visual attention bias during observation. Furthermore, we propose a 6DoF mesh saliency detection algorithm based on the uniqueness measure and the bias preference. To evaluate the proposed approach, we also design an evaluation metric accordingly which takes the 6DoF information into consideration, and extend some state-of-the-art 3D saliency detection methods to make comparisons. The experimental results demonstrate the superior performance of our approach for 6DoF mesh saliency detection, in addition to providing benchmarks for the presented 6DoF mesh saliency database. The database and the corresponding algorithms will be made publicly available for research purposes.
\end{abstract}
\vspace{0.35cm}

  \end{@twocolumnfalse} 
] 



\section{Introduction}
{\let\thefootnote\relax\footnote{{This work was supported by the National Natural Science Foundation of China under Grant 61771348. (Corresponding author: Zhenzhong Chen, E-mail:  \texttt{zzchen@ieee.org})}}}Visual attention mechanism plays an important role in the human visual system (HVS). It selects and extracts a subset of visual information before further processing in the brain to help the visual system interpreting complex scenes in real-time \cite{730558}. Analyzing the visual attention behaviour of humans is important for understanding and simulating the HVS. However, computationally identifying the visually salient regions that match the HVS is quite challenging since a huge amount of visual information floods into the brain at the same time. Accurately detecting salient regions is highly desirable \cite{6871397} for many perceptually based computer vision and graphics applications such as object detection \cite{liu2020deep}, quality assessment \cite{8742588} and video compression \cite{6942210}. 

During the past years, 3D meshes are becoming more convenient to obtain due to the rapid development of 3D acquisition technology. It has been widely used in many different fields in our daily life such as architectural design, city modeling, and computer simulation. Recently, many studies have been proposed which focused on saliency detection for 3D meshes. Different from saliency detection for traditional 2D data (RGB images and videos), 3D mesh contains depth information and usually consists of larger data size, making 3D mesh saliency detection a more challenging yet important research topic. Recently, many algorithms for mesh saliency detection have been proposed, for example, Leifman \textit{et al.} \cite{7393832} proposed an algorithm which calculates mesh saliency in a multi-scale manner and takes the foci of attention into consideration. Song \textit{et al.} \cite{song2014mesh} proposed a mesh saliency detection method which uses the properties of the mesh's log-Laplacian spectrum to capture saliency in the frequency domain and later localize salient regions in the spatial domain. 

3D meshes can be observed from any position and orientation in 3D space. Different viewing directions will lead to different regions of the mesh to be visualized. To better observe a 3D mesh, humans will change their viewing position and direction frequently to make their Field of View (FoV) focus on particular regions to obtain more detailed visual information. Consequently, the head movement (position and viewing direction) plays an important role in distributing human visual attention towards 3D meshes since it determines which part of the 3D mesh is visible to the HVS. Although existing 3D mesh saliency detection algorithms mentioned above have achieved plausible results, they are detecting mesh saliency in a `static' manner \cite{doi:10.1111} which does not take the influence of the subject's position and viewing direction (subject's 6DoF data) into consideration. The `static' mesh saliency detection algorithm can hardly be applied to applications that need `dynamic' mesh saliency detection, such as dynamic point cloud compression \cite{7405340} and interaction \cite{40774762}. In these applications, 3D meshes are processed under different conditions at each time, so the saliency detection results change over time according to different conditions. Due to the lack of research in `dynamic' mesh saliency detection, investigations about `dynamic' mesh saliency detection algorithm which uses the subject's 6DoF information are in an urgent need. 

To explore visual saliency for 3D meshes, a validation with respect to human performance is needed \cite{10.1145}. The eye-tracking experiment remains the main way for understanding human visual behaviour while observing 3D meshes \cite{doi:10.1111}. During the eye-tracking experiment, successive eye movements can be recorded which describe the pattern of visual attention \cite{8918321}. The successive eye movements are comprised of a series of fixations and saccades \cite{10.2307/1731012}. The fixations are the pauses over informative regions while the saccades are rapid eye movements between fixations \cite{Salvucci2000Identifying}. By analyzing the eye-movement data, the human's visual and cognitive process can be better explored. Although studies such as \cite{song2014mesh,7393832,10.1145} have been proposed which collect the eye movement data while subjects are observing 3D meshes, some of them just presented limited viewing directions for subjects to observe 3D meshes while others just displayed the rendered image of 3D meshes to subjects. Besides, all of these studies required the subjects to fix their heads at approximately the same position, which is impractical in real situations and might collect unreliable eye-movement data for 3D meshes. 

\begin{figure}[!t]
	\centering
	\includegraphics[width=8.0cm]{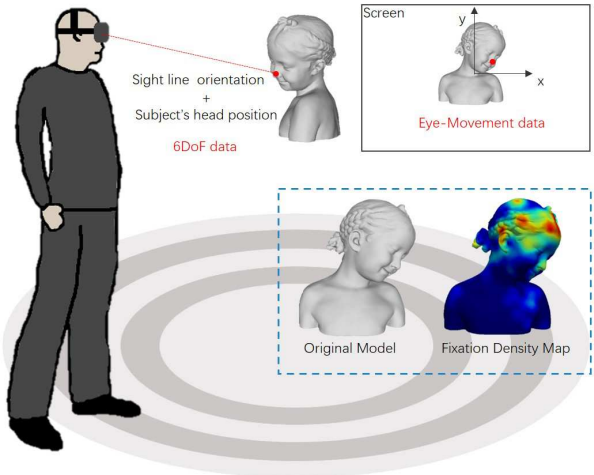}
	\caption{Illustration of the experiment. The red points represent the fixation points. The screen denotes the HTC Vive headset's screen. The left one in the blue dotted box is the original 3D mesh while the right one represents the collected fixation density map (FDM) for the mesh using 6DoF data. Blue/red colour suggests lower/higher fixation density value.}
	\label{fig1}
\end{figure}

Benefiting from the recent development of virtual reality and eye-tracking technologies, collecting  6DoF data of the subjects' heads and their eye-movement data simultaneously while observing 3D meshes are available. In this paper, a novel 6DoF mesh saliency database is developed by conducting an eye-tracking experiment for 3D meshes in the virtual reality environment. The objective of the proposed database is to provide more experimental data for exploring human visual attention behaviour while observing 3D meshes and facilitate more studies related to 6DoF mesh saliency detection. During the experiment, the HTC Vive headset is used for presenting 3D meshes to subjects with an immersive experience in the virtual scene and collecting the 6DoF data of the subjects' heads. The embedded aGlass eye-tracking module is capable of capturing eye-movement data when subjects are moving freely to observe 3D meshes. By combining the headset with the eye-tracking module, subjects can move freely in the virtual scene to focus their attention on visually salient regions on 3D meshes. Using the collected 6DoF data and eye-movement data, the fixation points of all subjects on each 3D mesh can be obtained by calculating the intersection points of the subject's actual sight-lines with the 3D mesh and applying fixation classification operations. A clear illustration of the experiment is shown in Fig. \ref{fig1}.

\begin{table*}[!t]
	\caption{Overview of mesh saliency databases. }
	\begin{threeparttable}
		\scalebox{0.9}{
			\begin{tabular}{llllll}
				\hline
				\textbf{Name} & \textbf{Size(Ages)}  & \textbf{Movement} &  \textbf{Visual Stimuli} & \textbf{Viewing Direction} &\textbf{Source Data} \\ \hline  \hline
				Wang \textit{et al.} \cite{7478427}   & 30 (19 $\sim$ 31) & Not Allowed & printed meshes & 1 fixed direction& eye-movement data\\
				Wang \textit{et al.} \cite{3272127}    & 70 (avg. 24)& Not Allowed & printed meshes & 7 fixed directions& eye-movement data\\
				Lavoue \textit{et al.} \cite{doi:10.1111}   & 20 (20 $\sim$ 40) & Not Allowed &  rendered image/video  & fixed rendering directions& eye-movement data\\
				\textbf{6DoFMS}     & 21 (20 $\sim$ 30) & Free Movement&  virtual meshes & free viewing direction& 6DoF $+$ eye-movement data \\ \hline
		\end{tabular}	}
		\footnotesize{Size (Ages): Number of people in the experiment and their age range. Movement: Head movement and body movement during the experiment. Source Data: Types of data collected in the database.}\\
	\end{threeparttable}
	\label{tabel1}
\end{table*}

To the best of our knowledge, there exists no database for 3D meshes which provides both the subject's 6DoF data and eye-movement data. Besides, due to the lack of the database, no 6DoF mesh saliency detection algorithm and evaluation metric has been proposed. The closest work is visual saliency databases for 3D meshes. Earlier work was proposed by Wang \textit{et al.} {7478427}, in which an eye-tracking experiment is conducted for 3D meshes to validate whether the saliency found in flat stimuli can be related to the 3D scene. Since their previous experiment \cite{7478427} is limited in the variation of viewing conditions towards 3D meshes, Wang \textit{et al.} \cite{3272127} later proposed another large database which collects human fixations on 3D meshes . In their experiment, 3D meshes are presented in several different viewing conditions such as different viewing directions and different materials. According to their experiment, they found that the salient feature seems to be related to semantically meaningful region on 3D meshes. Besides these studies, Lavoue \textit{et al.} \cite{doi:10.1111} also presented a database which records the eye-movement data for rendered 3D shapes. During their experiment, 3D meshes are rendered using different materials and lighting conditions under different scenes.

Although these databases have been proposed, our work differs from theirs in three aspects. 1) Visual stimuli. In \cite{3272127,7478427}, the 3D meshes are presented to subjects with limited viewing directions while in \cite{doi:10.1111}, the rendered videos of 3D meshes are shown on the screen for subjects to observe. Different from these visual stimuli, in the proposed database, the 3D meshes are presented in the virtual environment and subjects are allowed to observe them in any viewing direction. 2) Head movement. The databases proposed in \cite{3272127, 7478427, doi:10.1111} all require the subjects' heads to be fixed at a certain position during the experiment, while in the proposed database, subjects can move freely without limitations. 3) Source data. The databases proposed in \cite{3272127, 7478427, doi:10.1111} provide only the eye-movement data while the proposed database provides not only the eye-movement data but also the subjects' 6DoF data. An overview of these databases is provided in Table. \ref{tabel1}. Note that the proposed database is denoted as 6DoFMS.

The main contributions of this paper are summarized below: 
\begin{itemize}
	\item {A setup for 3D eye-tracking experiment in 6DoF is developed which allows the subject's free movement during mesh observation. The setup includes the intersection points calculation, fixation classification and fixation clustering procedure to obtain fixation points on 3D meshes.
	}
	
	\item {The first 6DoF mesh saliency database is developed which provides the 6DoF data and eye-movement data of 21 subjects on 32 3D meshes. The database can be used to explore visual attention behaviour in 6DoF and boost the development of mesh saliency detection in 6DoF.
	}
	
	\item {Further investigations based on the developed database are provided and behaviours of subjects' visual attention are analyzed, including the statistical characteristics of the database, the inter-observer variation, the viewing direction dependence and the visual attention bias.}
	
	\item {A novel algorithm for 6DoF mesh saliency detection is proposed which considers the uniqueness of the visible points and the bias preference using the 6DoF data. Based on the proposed 6DoF saliency evaluation metric, the algorithm outperforms 5 state-of-the-art 3D saliency detection methods. The experimental results will also provide benchmarks for the presented 6DoF mesh saliency database.}
\end{itemize}

The rest of this paper is organized as follows. Section 2 introduces the related work of 3D mesh saliency detection algorithms and eye-tracking experiments. Section 3 provides the details of the experiment and the data acquisition process. In Section 4, statistical analysis and further investigations toward the database are provided. In Section 5, a new 6DoF mesh saliency detection algorithm is proposed and an ablation study is conducted. In section 6, a 6DoF saliency evaluation metric is provided to evaluate the performance of the proposed approach. Finally, conclusions are summarized in Section 7. 
	
\section{Related Work}
In this section, we first summarize the related work about saliency detection. Then we review some work related to 3D visual saliency database.
\subsection{Saliency Detection}
Visual saliency helps the brain to obtain sufficient visual information and decide the attention distribution for a given scene \cite{Koch1985Shifts}. Visual saliency detection simulates the attention mechanism in the HVS \cite{1658355} and is of vital importance to many human-centered applications. It has been widely studied for decades and many algorithms have been proposed for both 2D and 3D data.

Various algorithms have been proposed to solve the 2D saliency detection tasks \cite{li2014visual, qin2018hierarchical, li2010probabilistic}. Early studies mainly focus on utilizing low-level features such as center-bias, contrast prior and background prior \cite{8578428} to highlight salient regions that are different from their surroundings. The earliest approach for image saliency detection was proposed by Itti \textit{et al.} \cite{730558}, in which colour, intensity, and orientation features are taken into consideration. Their method utilizes the center-surround operation to simulate the visual receptive fields and combines different saliency maps together to obtain the final saliency detection result. Later, several advanced approaches have been proposed for both image \cite{he2015supercnn} and video data \cite{souly2016visual}, for example, Wei \textit{et al.} \cite{978-3-642-33712-3_3} took the boundary prior and connectivity prior into consideration and proposed a geodesic saliency detection method for images. Kim \textit{et al.} \cite{7091884} proposed a random walk with restart based video saliency detection algorithm which can detect the foreground salient objects and suppress the background effectively. Recently, many deep learning algorithms have been proposed for saliency detection such as the work proposed by Zeng \textit{et al.} \cite{8578275}, in which a deep neural network is used to promote the precision of the captured image-specific saliency cues to achieve better saliency detection results. 

With the growing amount of 3D meshes, 3D saliency detection is in an urgent need. Early 3D saliency detection algorithms usually detect salient regions by calculating saliency values in its 2D projection. For example, Yee \textit{et al.} \cite{10.383748} projected the 3D dynamic scene into a reference image and then computed the saliency map to reduce the computational load for calculating the global illumination solution. However, the projection process leaves out the depth information of the 3D mesh and might lead to unsatisfying saliency detection results. Recently, many 3D mesh saliency detection algorithms have been proposed which are inspired by work of 2D saliency detection methods \cite{730558}. For instance, Lee \textit{et al.} \cite{1073204.1073244} proposed a novel mesh saliency detection algorithm  which uses the center-surround operator on Gaussian weighted mean curvatures and achieves more visually pleasing results. Wu \textit{et al.} \cite{WU2013255} took both local contrast and global rarity into consideration and presented a new approach for computing saliency values for 3D meshes. In their method, a multi-scale local shape descriptor is introduced to calculate the local contrast and the global rarity is defined by its speciality to all other vertices. Later, a linear combination is used to integrate both cues to obtain the final saliency detection result.

Besides processing 3D mesh data, several algorithms have also been proposed to deal with point cloud data. Shtrom \textit{et al.} \cite{6751558} presented an algorithm which can cope with extremely large point cloud data. Their algorithm takes the low-level distinctness, point association, and high-level distinctness into consideration and then combines these cues to obtain the final detection results. Tasse \textit{et al.} \cite{7410384} provided a cluster-based approach which decomposes the point cloud into several clusters using fuzzy clustering and evaluates the cluster uniqueness and spatial distribution to predict saliency value for the point cloud. Guo \textit{et al.} \cite{Yu2018Point} proposed a novel local shape descriptor based on covariance matrices to help point cloud saliency detection, and Yun \textit{et al.} \cite{7533123} came up with an algorithm which voxelizes the point cloud into supervoxels and then estimates the cluster saliency by computing the distinctness of geometric and colour features based on center-surround contrast. 

Although many 3D saliency detection algorithms have been proposed, they are detecting 3D visual saliency in a `static' manner without considering the subject's movement and can hardly be used to `dynamic' applications. Developing a `dynamic' 3D saliency detection method using 6DoF data is in an urgent need. 

\subsection{3D Visual Saliency Database}
To evaluate the effectiveness of the 3D saliency detection algorithms, a ground-truth database with either fixation point locations and fixation maps are needed. However, due to the limitations of the eye-tracking technologies, early studies were proposed without using eye-tracking device but using mouse tracking as an alternative interaction with subjects. Dutagaci \textit{et al.} \cite{DutagaciEvaluation} presented a benchmark which helps to evaluate the performance of 3D interest point detection algorithms and provided analysis using the subjective experimental data. During the experiment, users are allowed to freely rotate the 3D mesh and are asked to mark the interest points on it. Later, the ground-truth is constructed by utilizing a voting-based method and three evaluation metrics are presented to help the evaluation. Chen \textit{et al.} \cite{ChenSchelling} proposed a `Schelling point' mesh database by asking subjects to select points on 3D meshes that they think the others will select and a thorough analysis is proposed by analyzing the collected data to help designing mesh saliency detection algorithms. Besides, Lau \textit{et al.} \cite{Lau2016Tactile} also proposed a novel tactile mesh saliency database which helps to detect regions that human is more likely to touch and presented an algorithm which helps to measure tactile saliency. 

Recently, more 3D visual saliency databases have been proposed which use eye-tracking devices to collect the subject's eye-movement data. For example, Howlett \textit{et al.} \cite{10.1145/1077399.1077406} presented a database which captures human gaze data in 3D rendered images and examined whether the saliency information can help to improve the visual fidelity of simplified 3D meshes. Kim \textit{et al.} \cite{10.1145} compared previous 3D saliency detection results to eye movement data collected through an eye-tracking based user study and demonstrated that current saliency detection algorithms achieve better performance than curvature-based methods. Wang \textit{et al.} \cite{7478427} presented a database which contains 15 3D meshes and provided investigations about the human's fixation pattern while exploring 3D meshes. Later, Wang \textit{et al.} \cite{3272127} proposed another database which consists of 16 3D meshes. In this database, different kinds of low-level and high-level features such as smooth surface and task-related affordances are taken into consideration. Different from Wang's databases \cite{7478427, 3272127} which present the printed 3D meshes to subjects, Lavoue \textit{et al.} \cite{doi:10.1111} proposed a database which collects the visual saliency by presenting the rendered 3D meshes to subjects. During the experiment, influencing factors such as illumination and materials are studied using the collected data. 

The experiments conducted by Dutagaci \textit{et al.} \cite{DutagaciEvaluation} and Chen \textit{et al.} \cite{ChenSchelling} used the mouse-tracking technologies, thus they might be inappropriate for exploring visual saliency on 3D meshes. However, other approaches like \cite{7478427, 3272127, doi:10.1111} are collected under several constraints which might influence the subject's real observing behaviour, such as requiring the subject's head to be fixed at a certain position during the experiment.  More importantly, none of these databases collect 6DoF information, thus they can hardly be used for building and evaluating 6DoF mesh saliency detection algorithm. To address this problem, a novel 6DoF mesh saliency database is presented which uses the HTC Vive headset with aGlass eye-tracking module to collect both the 6DoF data and eye-movement data.

\section{Data Acquisition}
The main idea of the proposed experiment is to capture subject's eye-movement data on 3D meshes and record their 6DoF data. Following the established protocols of the eye-tracking experiment \cite{GOTTLIEB2013585}, our experiment can be categorized into three steps. During the first step, calibration is presented to each subject to ensure the accuracy of the eye-movement data. Then, 3D meshes are presented in the virtual scene using the HTC Vive headset for subjects to observe. During the observation, subject's 6DoF data and eye movement data are recorded simultaneously. Later, fixation points are detected using both the 6DoF data and eye-movement data for further investigation. So in this section, we will first illustrate the design and setup of our experiment and then describe the procedure of data processing in detail. A brief summary of the experimental setup is shown in Table. \ref{table2}.

\begin{table}[!t]
	\caption{Experimental details of the proposed database.}
	\centering
	\scalebox{0.85}{
	\begin{tabular}{lll}
		\hline
		\multicolumn{1}{l}{ \bfseries{Category}} & \bfseries{Details} & \bfseries{6DoFMS}  \\
		\hline \hline
		\multirow{1}{*}{Participant} & Size (Ages) &21(20 $\sim$ 30) \\
		\multirow{1}{*}{} & Male/Female &11/10 \\
		\multirow{1}{*}{Condition} & Environment & quiet test room \\
		\multirow{1}{*}{} & Illumination & 3 spot-lights from different directions \\
		\multirow{1}{*}{} & Task & free viewing with no particular task \\
		\multirow{1}{*}{Device} & HTC Vive &present mesh and record 6DoF data   \\
		\multirow{1}{*}{} & aGlass  &record eye-movement data   \\    
		\multirow{1}{*}{Stimulus} & Duration & 10s for each mesh  \\
		\hline
	\end{tabular} }
	\label{table2}
\end{table}

\subsection{Design of the Experiment}

\noindent \textbf{Creation of the Stimuli:} 
It has been widely acknowledged that low-level features and high-level information will influence human visual behaviour \cite{GOTTLIEB2013585}. To better explore human visual behaviour for 3D meshes, we take the low-level features generated by 3D geometry and high-level information embedded in 3D shapes into consideration. Considering that the database proposed by Lavoue \textit{et al.} \cite{doi:10.1111} contains more 3D meshes when compared with the other databases, we decided to use the same 3D meshes as in their database in our experiment to make a better comparison. These 32 3D meshes consist of four different types, including Humans, Animals, Familiar Objects and Mechanical Parts, each type conveys varying semantic information. 

\hspace*{\fill} \\
\noindent \textbf{Presentations:} Studies have been proposed which prove that the illumination and object material will have an impact on human visual behaviour \cite{doi:10.1111}. To avoid the possible distractions caused by illumination and object material, the 3D meshes used in the experiment are generated with the same material and presented under the same lighting condition. During the experiment, each 3D mesh is presented in the center of the virtual scene (a virtual room with white walls) with three spot-lights illuminating from three different directions and each 3D mesh is presented for 10s. Note that each 3D mesh is presented to each subject only once to avoid potential confounding effects of habituation or boredom for the repeated presentation of the same 3D mesh. 

Different from the experiment proposed by Wang \textit{et al.} \cite{3272127} which provides limited viewing directions and the experiment conducted by Lavoue \textit{et al.} \cite{doi:10.1111} which fixes the rendering direction of 3D meshes, in our experiment, subjects are allowed to move freely to observe the 3D mesh in any viewing direction. We believe this will help to collect more convincing 6DoF data and eye-movement data. 

\begin{figure*}[t]
	\begin{center}
		\subfigure[]{\includegraphics[width=5.5cm]{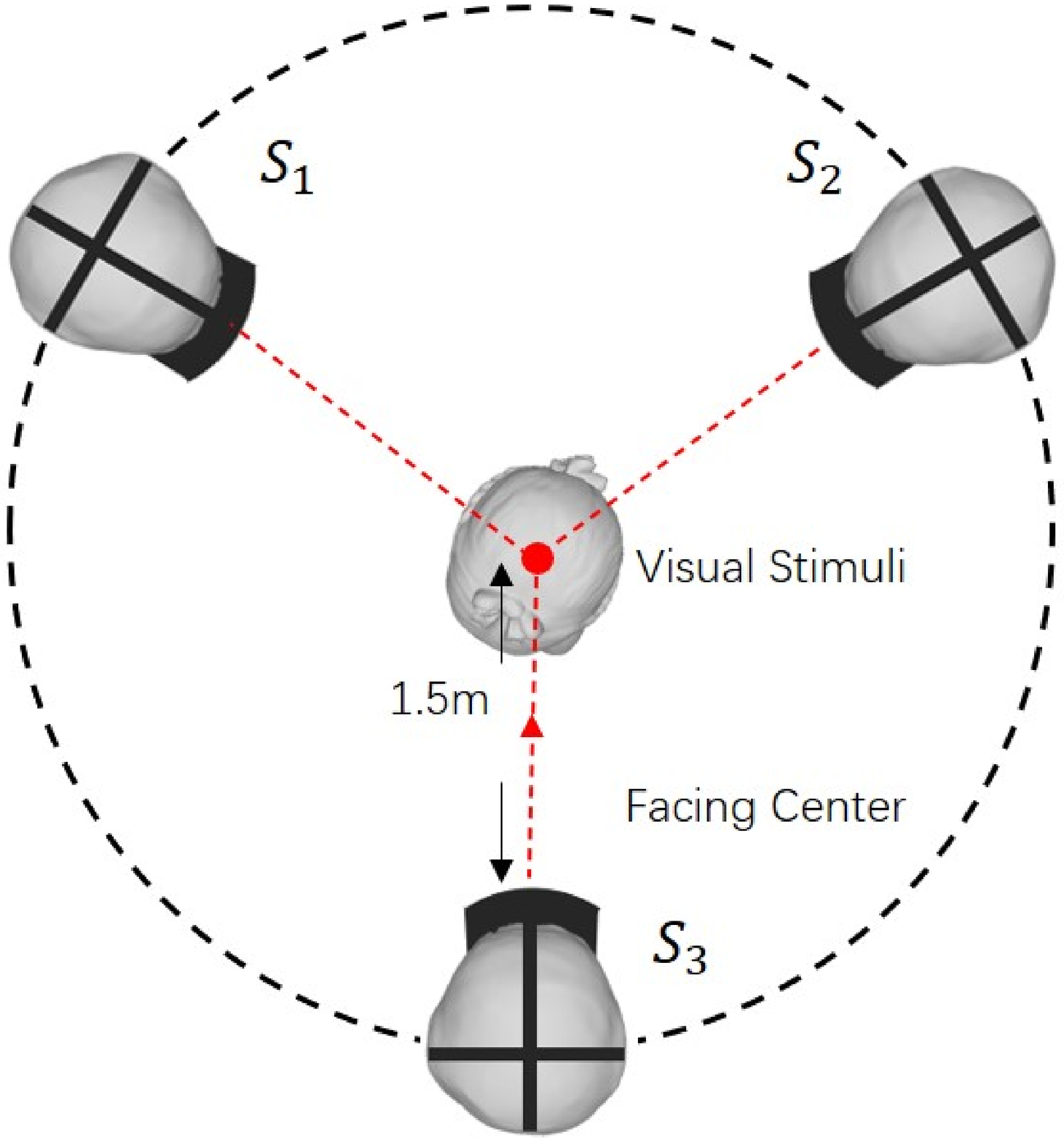}}
		\subfigure[]{\includegraphics[width=11cm]{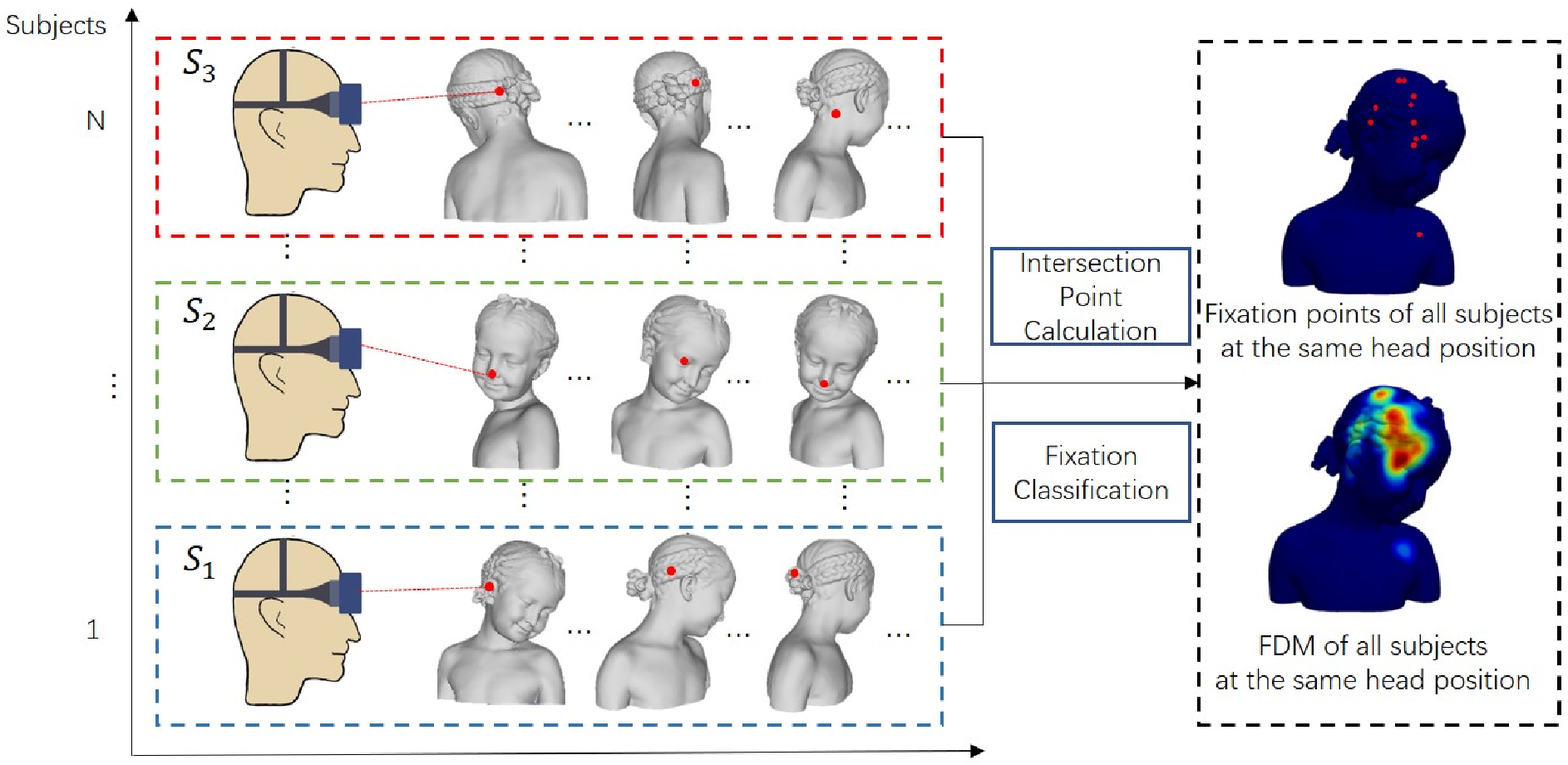}}		
	\end{center}
	\caption{Illustration of the experiment. (a) shows the distribution of the starting points and (b) is the flowchart of the experiment. The three heads in (a) represent the evenly distributed starting points while the red lines denote that subjects are asked to face the center of the virtual scene before each observation. The red, green and blue dotted boxes in (b) present the three test groups respectively. The top of the black dotted box shows the fixation points of all subjects collected at the same head position while the bottom denotes the corresponding Fixation density map (FDM).}
	\label{fig2}
\end{figure*}

\hspace*{\fill} \\
\noindent \textbf{Participants:} 21 subjects aged between 20 and 30 participated in the experiment, among them are 11 males and 10 females. They have normal or corrected to normal visual acuity and no (known) colour deficiencies. More importantly, all the subjects are naive about the purpose of the experiment and are told to observe all the 3D meshes freely without particular tasks.

\hspace*{\fill} \\
\noindent \textbf{Apparatus:} 
The HTC Vive headset is used to present the 3D meshes in the virtual scene and record the 6DoF data of the subject's head. The display resolution of the headset is 1080*1200 with a refresh rate of 90Hz. Every recorded 6DoF data is characterized by the coordinate of the virtual scene which includes the headset's position data and orientation data. 

In addition to the recorded 6DoF data, the eye-movement data is captured by the aGlass eye-tracking module which is embedded in the HTC Vive headset. The eye-tracking module is capable of capturing  eye-movement data within the FoV at less than $0.5^{\circ}$ error. When tracking the eye-movement data, the sampling rate is set to 120 Hz. Each recorded eye-movement data is characterized by headset's screen coordinate and will be later classified into fixations and saccades. 

\hspace*{\fill} \\
\noindent \textbf{Calibration Details:}
To ensure the accuracy of the recorded eye-movement data, a calibration step needs to be performed before each experimental section. Note that in this paper, we use the calibration software provided by the aGlass module. During the calibration step, 9 moving points will appear on the screen of the headset successively and subjects need to follow the movement of these calibration points. After the calibration, the software will compute the accuracy and precision, a verification process will also appear for subjects to check whether the calibration is within a tolerable error range. If the subject fails to pass the calibration, he/she will be asked to recalibrate to ensure the accuracy of the collected eye-movement data.

\hspace*{\fill} \\
\noindent \textbf{Procedure:} 
The experiment was conducted in a quiet room with each subject participating individually. An illustration of the experiment is shown in Fig. \ref{fig2}. Before the experiment, three starting points (Fig. \ref{fig2} (a)) $S_1$, $S_2$ and $S_3$ were distributed evenly in the test room. The distance between each starting point and the center of the virtual scene was set to 1.5m, since too close to the 3D mesh may fail to provide subjects an overall view of the 3D mesh  while too far away from the 3D mesh cannot provide more detailed information, and 1.5m provides a better observing experience for all the 3D meshes in our database. During the experiment, 21 subjects were randomly separated into three test groups (Fig. \ref{fig2} (b)) with each test group containing 7 subjects. Each test group was assigned a specific starting point $S_1$, $S_2$ or $S_3$ to begin their observation. Note that each starting point has a raised mark on the ground to help subjects verify that they have returned to their starting point after observing a 3D mesh.  According to the figure, subjects were asked to face the center of the room before observing each 3D mesh (shown as the red lines in Fig. \ref{fig2} (a)). The purpose of the starting points' design is to encourage subjects to observe the 3D meshes in all directions.

\begin{figure}[!t]
	\centering
	\includegraphics[width=8.0cm]{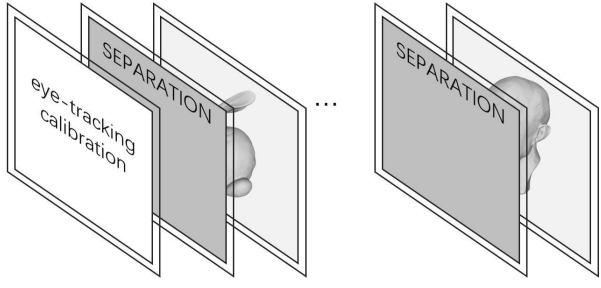}
	\caption{Example of an experimental section. At the beginning of each experimental section, an eye-tracking calibration is needed and between each section, subjects will have 2-minute to rest.}
	\label{fig3}
\end{figure}

During the experiment, 32 3D meshes were randomly divided into four sections with each section containing 8 meshes. Each mesh was presented for 10-second since presenting the mesh for longer time might lead to visual fatigue while shorter time might be insufficient for the observation. According to our experiment, we found that 10-second duration provides a better balance. A clearer illustration of an experimental section is depicted in Fig. \ref{fig3}. Note that a gray image was shown on the screen of the headset before observing each 3D mesh to leave enough time for subjects to return to their starting point and prepare for the next observation. To avoid visual fatigue and motion sickness, each subject had 2-minute rest after observing each experimental section of 8 meshes. 

Compared with other databases which limit the subject's head movement using a chin and forehead rest when observing 3D meshes \cite{3272127,7478427,doi:10.1111}, in our experiment, the only constraint for subjects is that they need to start from their starting point and face the center of the virtual scene before each observation. When the observation begins, subjects were allowed to move freely in the virtual scene to observe the 3D meshes without limitations and tasks. The distance between the subjects and the 3D mesh can be changed freely according to the subject's preferences. Besides, the subject's viewing direction was also allowed to change freely, which is different from other databases that provide limited or fixed viewing directions. We think this vivid virtual observing experience will make subjects feel that they are observing 3D meshes in the real world, helping us to collect more convincing and reliable 6DoF data and eye-movement data.

\subsection{Data Processing}
With the support of the HTC Vive headset and the embedded aGlass eye-tracking module, the subject's 6DoF data and eye-movement data can be recorded simultaneously. Note that the recorded data for the $i$-th subject on $j$-th mesh is denoted as $H^{ij}$. Each $H^{ij}$ is a $K*3$ matrix where $K$ represents the number of the collected 6DoF data on the mesh and $H_k^{ij}$ represents the $k$-th collected 6DoF data and eye-movement data of subject $i$ on mesh $j$. For each $H_k^{ij}$, $H_k^{ij} = \{ P_k^{ij},O_k^{ij},S_k^{ij}\}$, in which $P_k^{ij}$ represents the head position data, $O_k^{ij}$ is the head orientation data and $S_k^{ij}$ denotes the eye-movement data.  

Based on the collected data, the intersection points of the subject's actual sight-lines and the 3D mesh can be calculated and later used for fixation classification. The relevant symbols used for data processing and analysis are summarized in Table. \ref{tab:symbol}.

\begin{figure}[!t]
	\begin{center}
		\includegraphics[width=8.0cm]{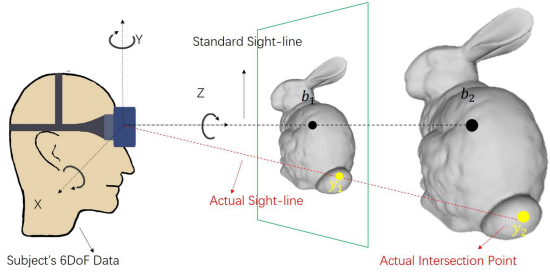}	
	\end{center}
	\caption{An illustration of the standard sight-line and the actual sight-line. The black line represents the standard sight-line while the red line denotes the actual sight-line. The green shape represents the screen of the HTC Vive headset. The black point $b_2$ denotes the intersection point of the standard sight-line with the mesh. The yellow point $y_2$ shows the actual intersection point of the actual sight-line with the mesh.}
	\label{fig4}
\end{figure}

\hspace*{\fill} \\
\noindent \textbf{Intersection Points Calculation:}
Before calculating the intersection point, the difference of the subject's head orientation and actual sight-line orientation needs to be clarified. The head orientation decides the orientation of the  standard sight-line representing that subjects are looking straight ahead, while the actual sight-line orientation represents the direction that the subject is actually looking.  When observing an object, subjects will first move their heads to make the important part of the mesh visible to them and this is recorded as the head orientation. Later, subjects tend to slightly move their sights up or down (left or right) instead of moving their heads to have a better observation of the details and this is the actual sight-line orientation. An illustration of the standard sight-line and the actual sight-line is shown in Fig. \ref{fig4} with the black dotted line denoting the standard sight-line and the red dotted line representing the actual sight-line. Note that the change from the standard sight-line to the actual sight-line is determined by the eye-movement data. In order to obtain the actual intersection point, the linear equation of the subject's actual sight-line needs to be calculated.

\begin{table}
	\centering
	\caption{Symbol Table}
	\label{tab:symbol}
	\scalebox{0.85}{
	\begin{tabular}{ll}
		\hline
		Symbol         & Explanation \\
		\hline
		$P_k^{ij}$        & the $k$-th head position for subject $i$ on mesh $j$ \\
		$O_k^{ij}$        & the $k$-th head orientation for subject $i$ on mesh $j$ \\
		$S_k^{ij}$        & the $k$-th eye-movement data for subject $i$ on mesh $j$ \\
		$\mathbb{O}_k^{ij}$  & the head orientation calculated using  $O_k^{ij}$   \\
		$Y_k^{ij},B_k^{ij}$ & the coordinate of the points shown in Fig. \ref {fig4} \\
		$I_k^{ij}$ & the $k$-th intersection point of subject $i$ on mesh $j$ \\
		$D_k^{ij}$ & the current distance of subject $i$ to mesh $j$.  \\
		$M_{i}^{j}$ & the fixation density map for subject $i$ on mesh $j$ \\
		$S_{i_1i_2}^{j_1j_2}$ & the similarity between FDM $M_{i_1}^{j_1}$ and $M_{i_2}^{j_2}$\\
		$V^{PO}$   & the visible points on the mesh using 6DoF data $(P,O)$ \\
		$F^{PO}$   & the fixation points on the mesh using 6DoF data $(P,O)$ \\
		$v_i$   & the $i$-th visible point in $V^{PO}$ \\
		$f_{ni}$ & the $n$-th bin of the FPFH descriptor for $v_i$ \\
		$G_w^j$      & the ground-truth FDM for the $w$-th 6DoF data on mesh $j$ \\
		$R_w^j$ & the saliency result for the $w$-th 6DoF data on mesh $j$ \\
		\hline
		\hline
	\end{tabular} }
\end{table}

To calculate the intersection points, the head orientation is first obtained using the formula below. Note that all the data are collected in the left-hand coordinate system:

\begin{equation}
\mathbb{O}_k^{ij} = {R_z}*{R_x}*{R_y}*vector
\end{equation}
where $\mathbb{O}_k^{ij}$ represents the head orientation and the $vector$ is $[0,0,1]$. $R_x$, $R_y$, and $R_z$ represent the rotation matrices calculated using the orientation data $O_k^{ij}$ respectively. 

With the obtained head orientation $\mathbb{O}_k^{ij}$ and head position $P_k^{ij}=(p_{kx}^{ij}, p_{ky}^{ij}, p_{kz}^{ij})$, the linear equation of the standard sight-line can be calculated as:

\begin{equation}
\label{Eq:2}
\frac{{x - {p_{kx}^{ij}}}}{{\mathbb{O}_k^{ij}(1)}} = \frac{{y - {p_{ky}^{ij}}}}{{\mathbb{O}_k^{ij}(2)}} = \frac{{z - {p_{kz}^{ij}}}}{{\mathbb{O}_k^{ij}(3)}}\\
\end{equation}
where $\mathbb{O}_k^{ij}(1)$, $\mathbb{O}_k^{ij}(2)$, $\mathbb{O}_k^{ij}(3)$ denote the parameters from $\mathbb{O}_k^{ij}$ respectively.

According to Fig. \ref{fig4}, the black point $b_1$ is the intersection point of the standard sight-line with the screen while the yellow point $y_1$ denotes the intersection point of the actual sight-line with the screen. In the experiment, the distance between the headset's screen and the subject's head is fixed, so the coordinate of the black point $b_1$ in the virtual scene  $B_k^{ij}=(b_{kx}^{ij}, b_{ky}^{ij}, b_{kz}^{ij})$ can be obtained using Eqn. \ref{Eq:2}.

To calculate the coordinate of the yellow point $y_1$ in the virtual scene $Y_k^{ij}=(y_{kx}^{ij}, y_{ky}^{ij}, y_{kz}^{ij})$, the eye-movement data $S_k^{ij}=(s_{kx}^{ij}, s_{ky}^{ij})$ is needed since it denotes the movement of the actual sight-line from the standard sight-line. In our experiment, $Y_k^{ij}$
is calculated using the formula below:

\begin{equation}
\begin{array}{l}
{y_{kx}^{ij}} = {b_{kx}^{ij}} + {s_{ky}^{ij}}*\cos \alpha *\cos \beta  + {s_{kx}^{ij}}*\sin \beta \\
{y_{ky}^{ij}} = {b_{ky}^{ij}} - {s_{ky}^{ij}}*\sin \alpha \\
{y_{kz}^{ij}} = {b_{kz}^{ij}} + {s_{ky}^{ij}}*\cos \alpha *\sin \beta  - {s_{kx}^{ij}}*\cos \beta 
\end{array}
\end{equation}
where $\alpha$ represents the angle between the standard sight-line and the positive direction of the $Y$ axis and $\beta$ denotes the angle between the projection of the standard sight line on the $XoZ$ plane and the positive direction of the $X$ axis.

With the obtained subject's head position $P_k^{ij}$ and the position of the yellow point $y_1$, the linear equation of the actual sight-line can be calculated as:

\begin{equation}
\label{Eq:4}
\frac{{x - {p_{kx}^{ij}}}}{{{y_{kx}^{ij}} - {p_{kx}^{ij}}}} = \frac{{y - {p_{ky}^{ij}}}}{{{y_{ky}^{ij}} - {p_{ky}^{ij}}}} = \frac{{z - {p_{kz}^{ij}}}}{{{y_{kz}^{ij}} - {p_{kz}^{ij}}}}
\end{equation}

Using Eqn. \ref{Eq:4} and the position of the 3D mesh, the intersection point of the actual sight-line with the 3D mesh (the yellow point $y_2$) can be calculated, which is denoted as $I_{k}^{ij}$.

\hspace*{\fill} \\
\noindent \textbf{Fixation Classification:}
Following the procedure mentioned above, the intersection point $I_{k}^{ij}$ can be obtained for every subject on all 3D meshes. These intersection points present the viewing behaviour of subjects and can be parsed into periods of fixation points and saccade points. The fixation brings the region of interest (ROI) onto the fovea where the visual acuity is at the maximum while the saccade is the rapid transition between fixations \cite{Salvucci2000Identifying}. To better explore the subject's viewing behaviour towards the 3D mesh, the saccades and fixations need to be classified.

Fixation classification has been studied for a long time and many algorithms have been proposed. Fixation classification algorithms can be roughly divided into three categories \cite{Salvucci2000Identifying}: velocity based, area based and dispersion based. Velocity based algorithm considers the fact that fixation points have lower velocities when compared with saccade points while the area based algorithm identifies points within given AOIs that represent the visual targets. Dispersion based algorithm is built under the assumption that fixation points are more likely to occur near one another. In our experiment, the velocity threshold based identification method (I-VT) is utilized \cite{NystrAn} since it classifies the points straightforwardly and has been proven to be mature, effective and of low time complexity.

The I-VT fixation classification algorithm calculates velocity as the simple distance between two consecutive points so that the temporal component can be ignored. Then a predefined velocity threshold is used to separate points into fixations and saccades. However, the threshold needs to be chosen carefully, otherwise, it might lead to unsatisfying classification results. 

\begin{figure}[!t]
	\centering
	\subfigure[]{\includegraphics[width=2cm]{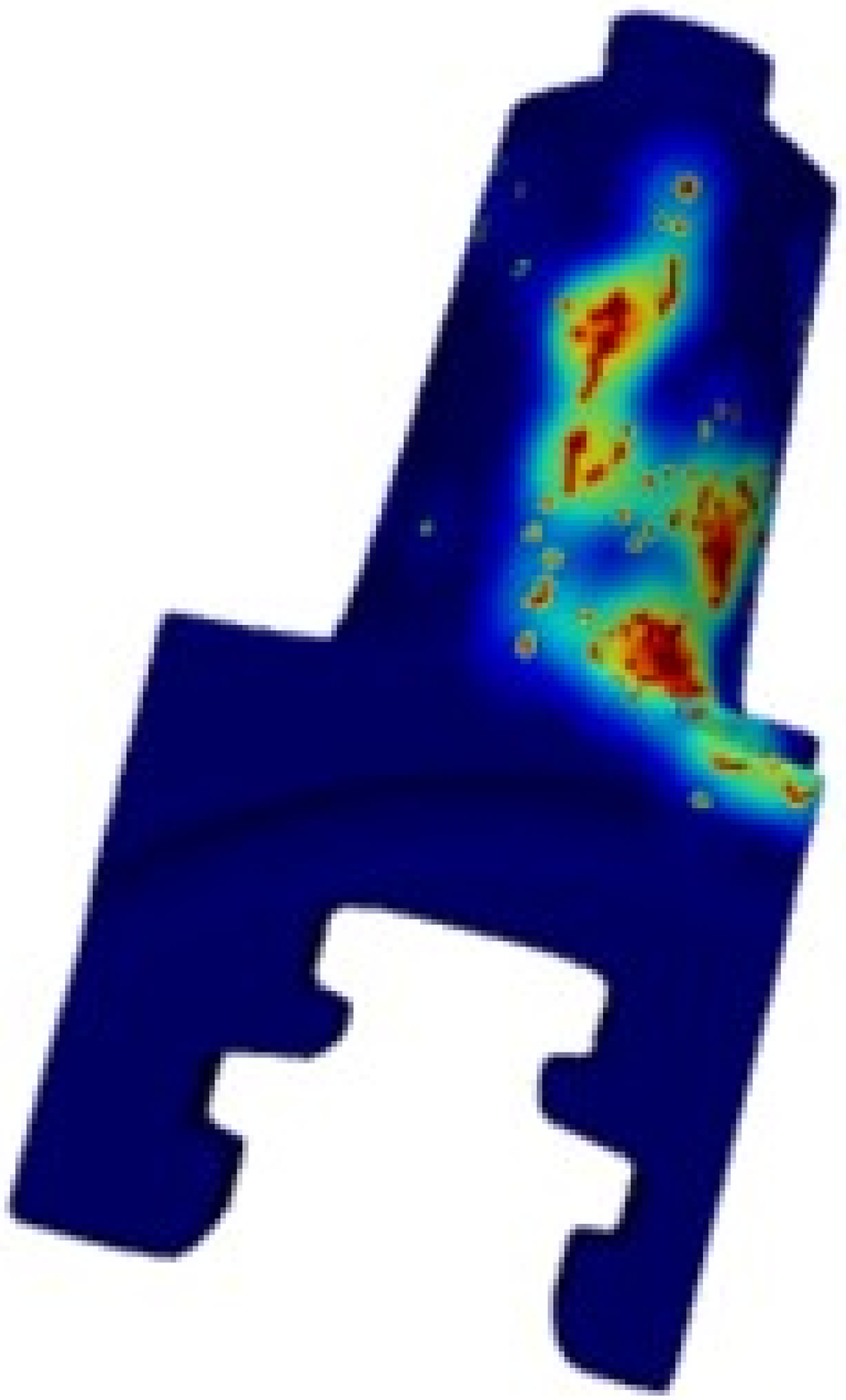}}
	\subfigure[]{\includegraphics[width=2cm]{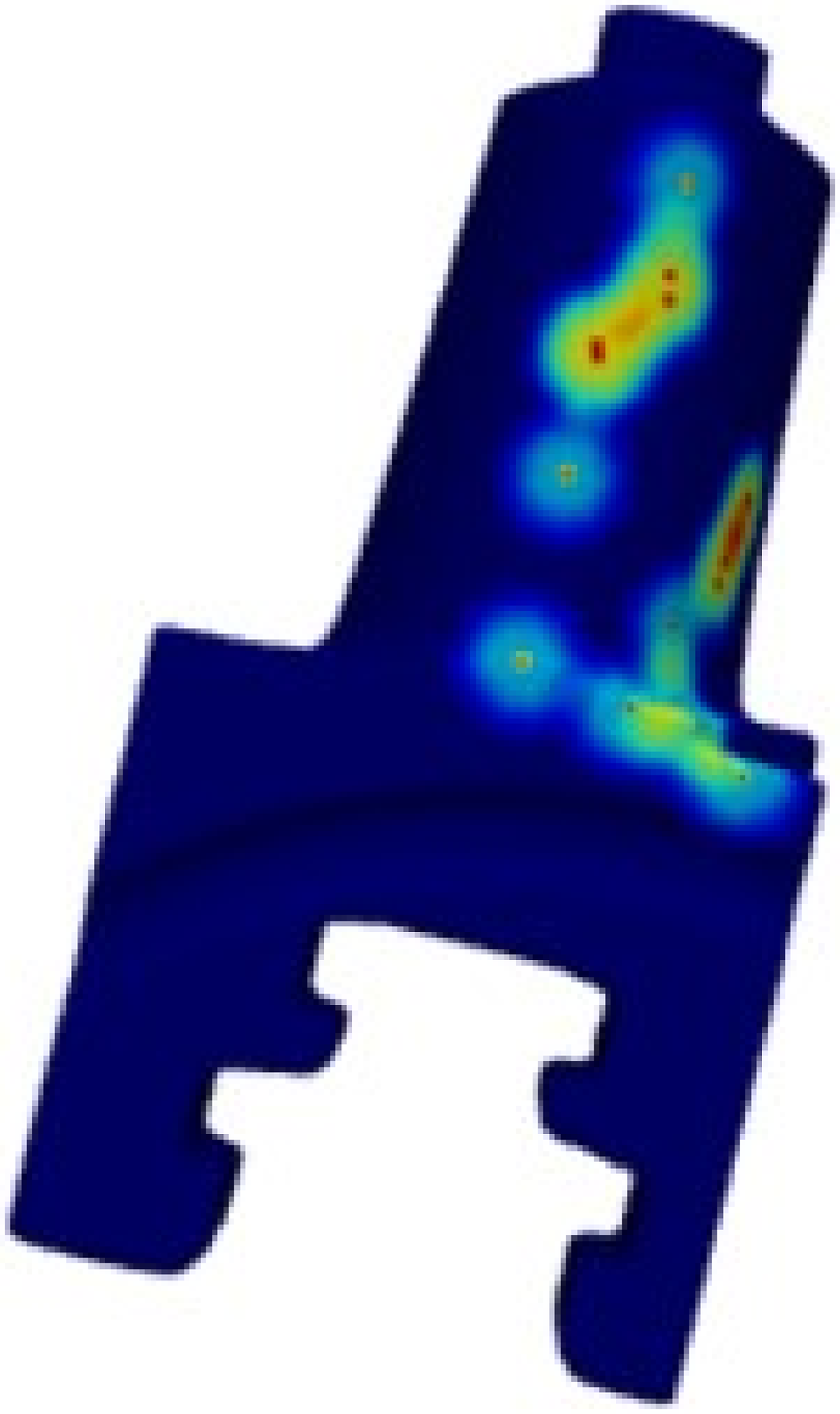}}
	\subfigure[]{\includegraphics[width=2cm]{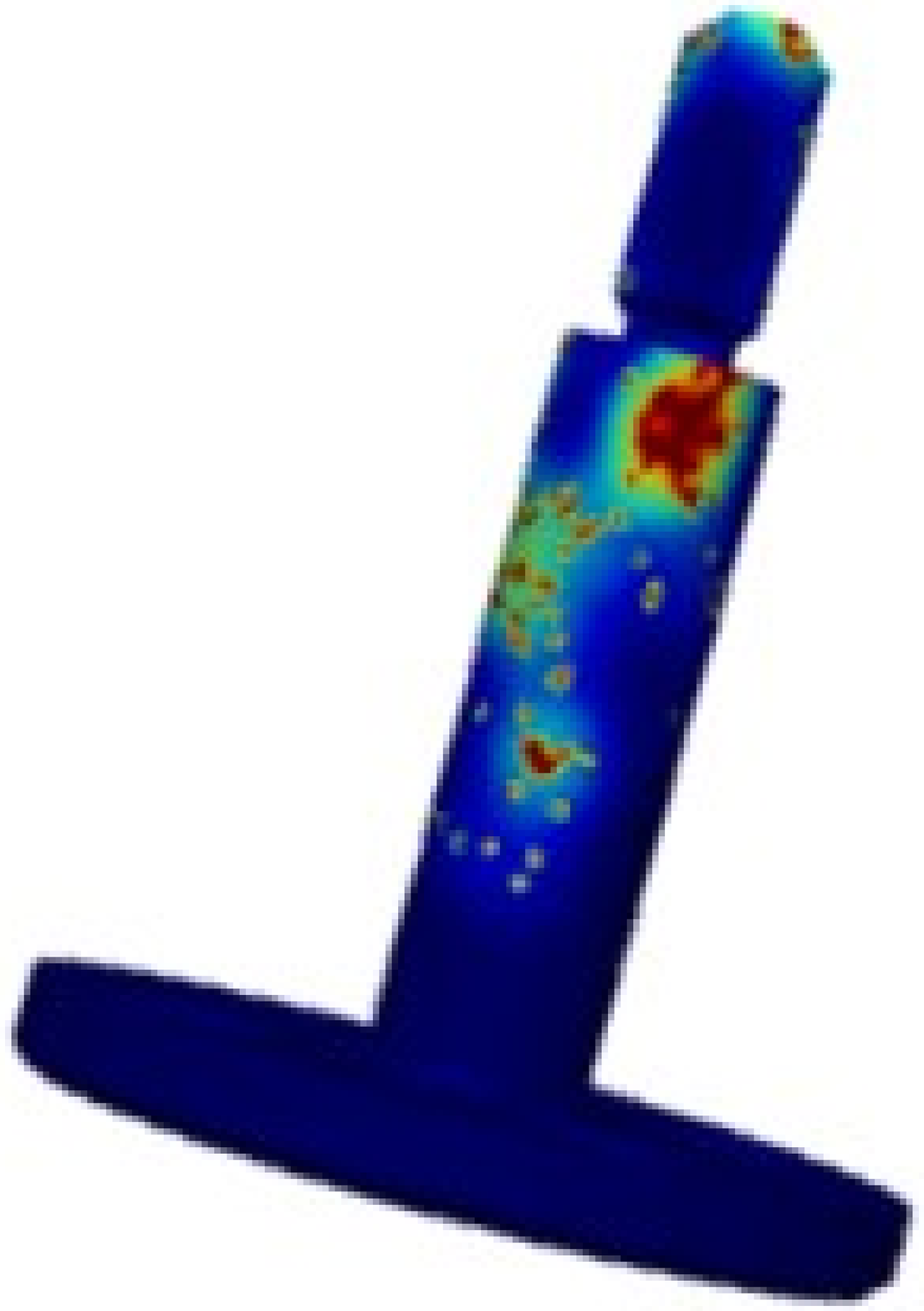}}
	\subfigure[]{\includegraphics[width=2cm]{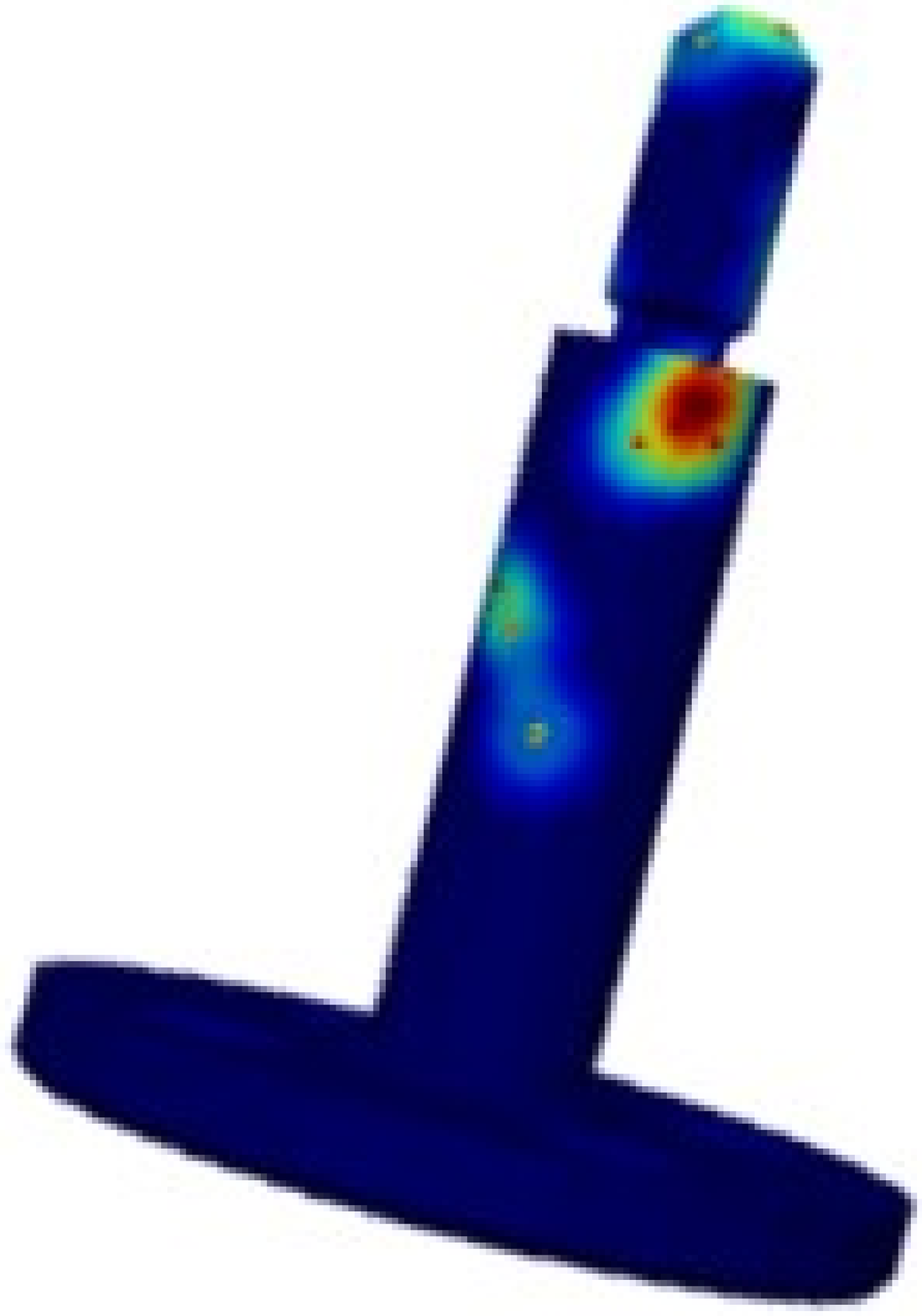}}
	\caption{Examples of the fixation clustering results. The blue/red region suggests the lower/higher fixation density. The red points in (a) and (c) are the original fixation points, while the red points in (b) and (d) denote the clustered fixation points.}
	\label{fig5}
\end{figure}

\begin{figure*}[!t]
	\begin{center}
		\subfigure[]{\includegraphics[width=6.5cm]{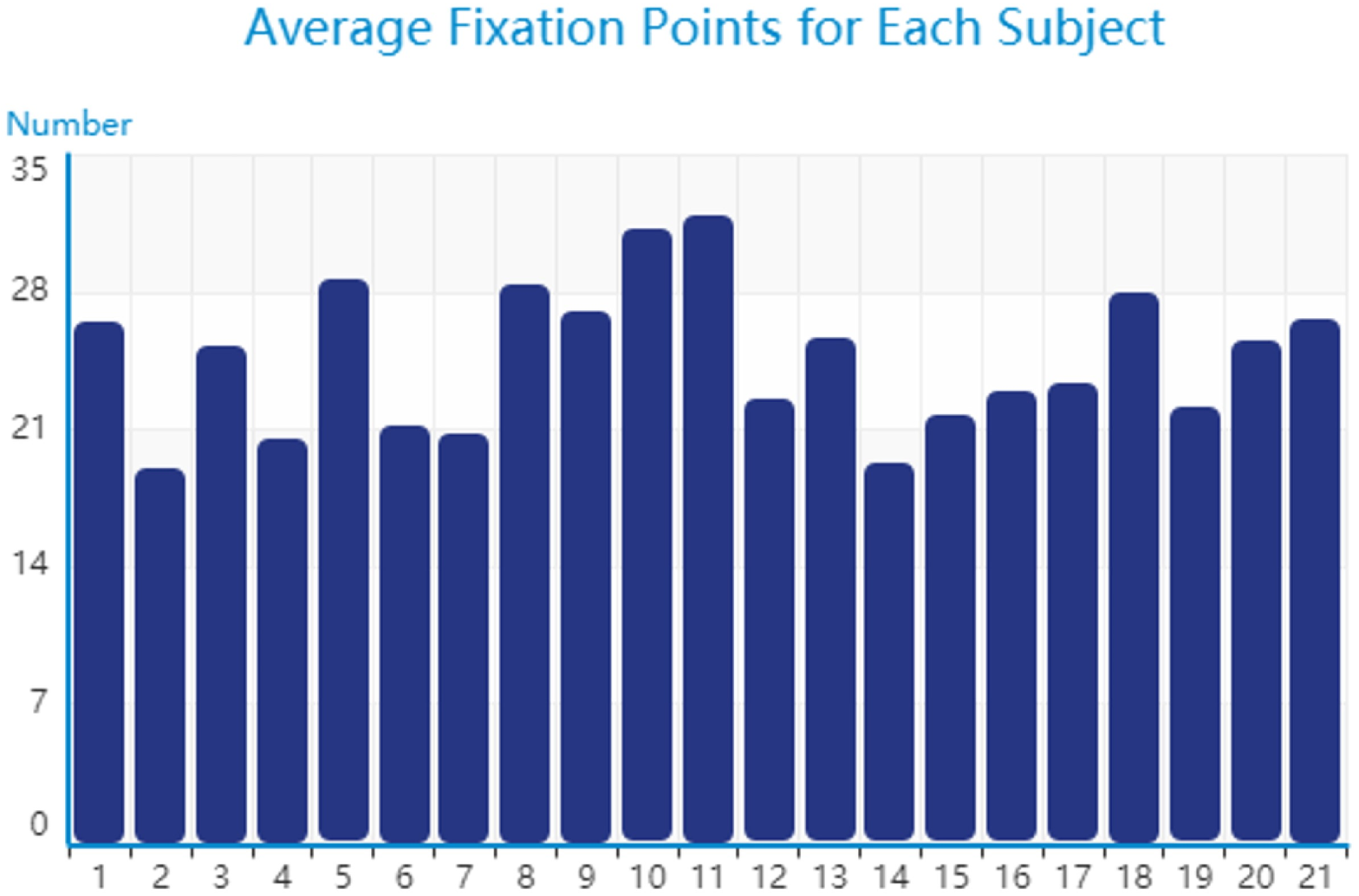}}
		\subfigure[]{\includegraphics[width=6.5cm]{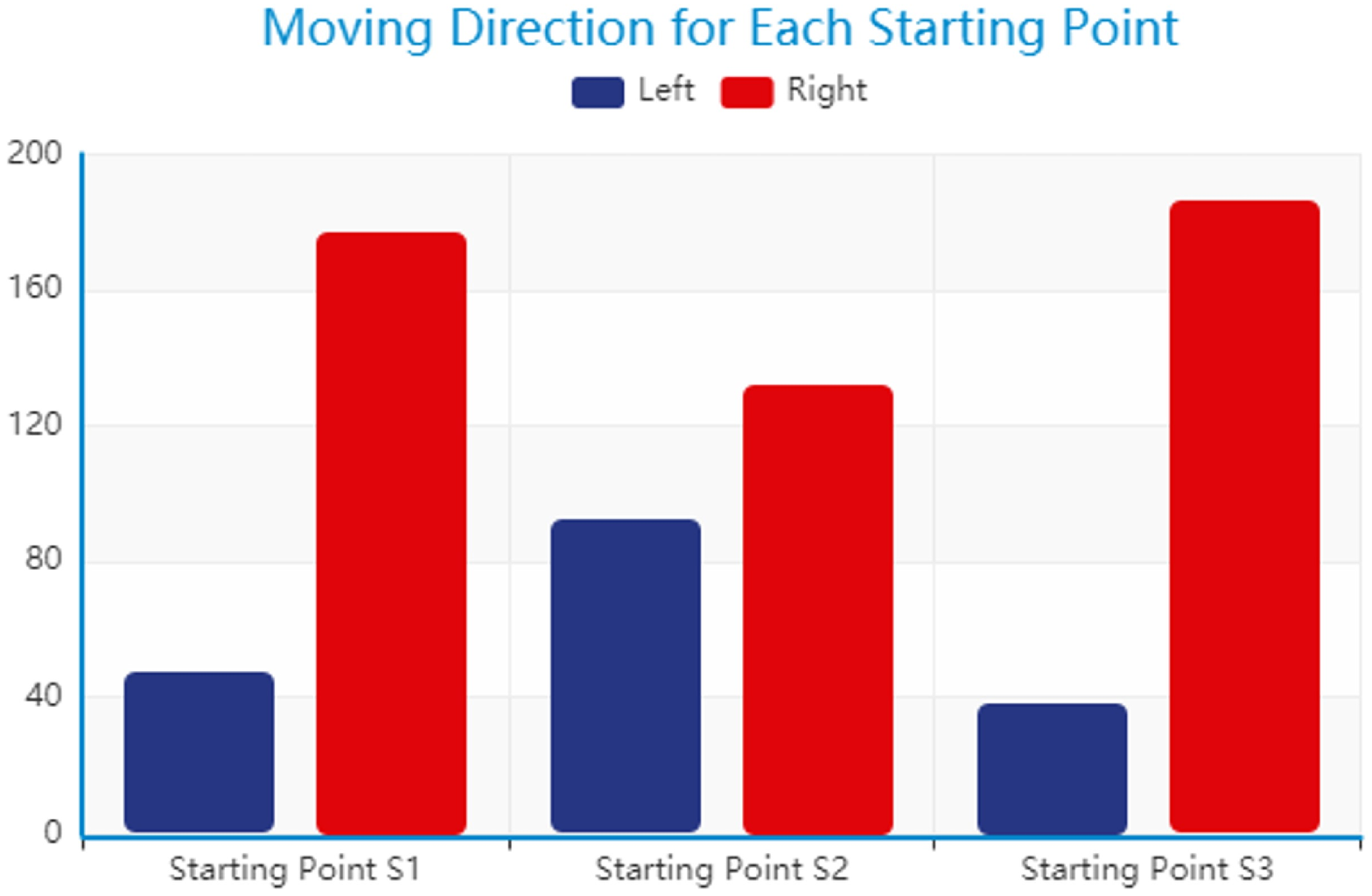}}
		\subfigure[]{\includegraphics[width=4.2cm]{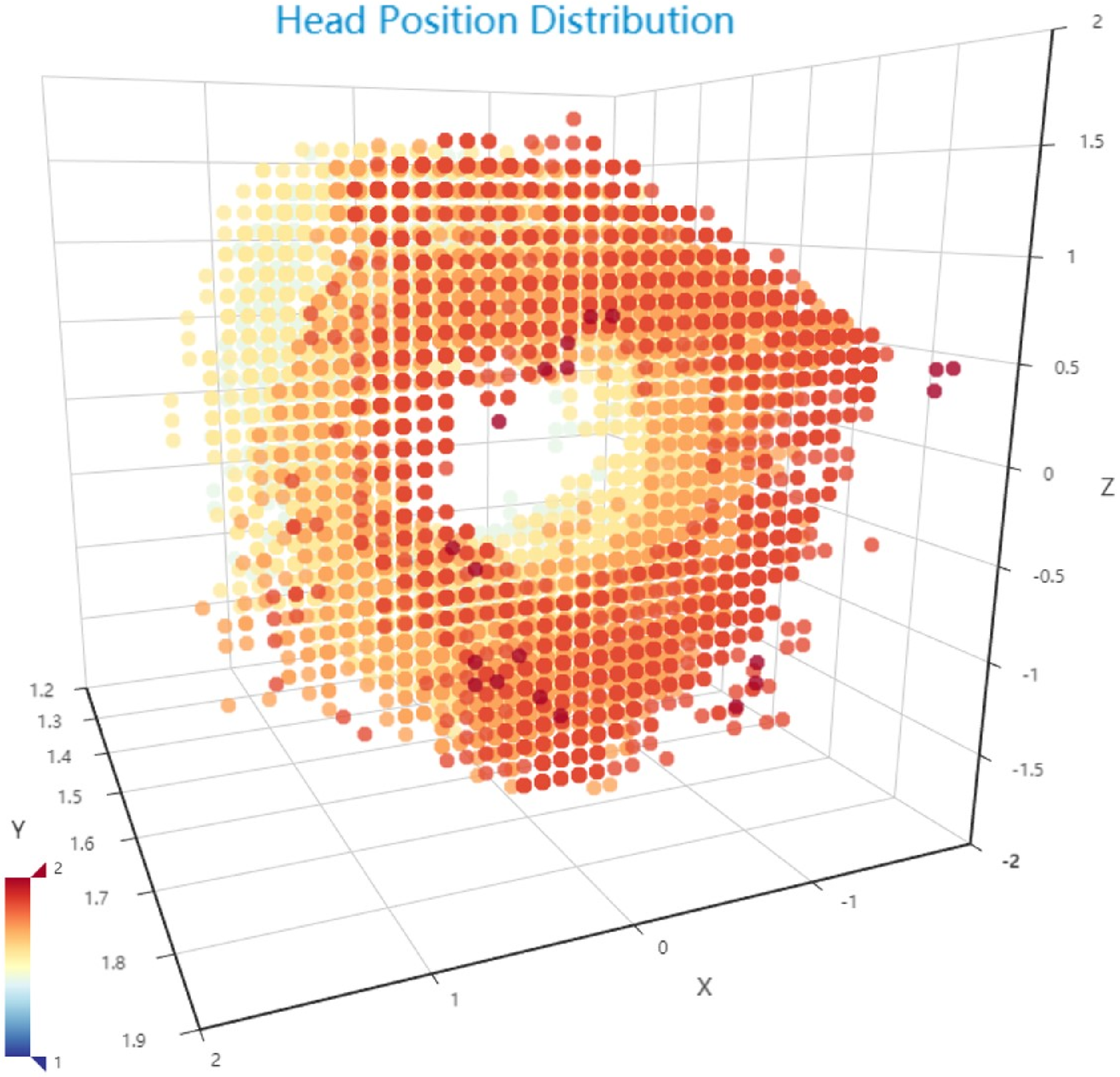}}
	\end{center}
	\caption{Statistical analysis results. (a) represents the average number of fixation points collected from each subject on all 32 3D meshes. (b) presents the frequency of subjects moving left or right when they start to observe the 3D meshes while (c) shows the distribution of the head position in the coordinate system.}
	\label{fig15}
\end{figure*}

When classifying fixation points for 2D images, the distance between the subject and the image is used fixed, an appropriate velocity threshold might work well for all the points in the image. However, a fixed threshold might not work well in our experiment since the distance between the subject and the mesh is always changing. For instance, if we move $1^{\circ}$ visual angle, the corresponding movement of the intersection point is approximately 0.3 cm on a mesh 50 cm away from the subject, but if the distance between the subject and the mesh expands to 1m, the movement of the point on the mesh will be approximately 0.6 cm. To address this problem, an adaptive velocity threshold is proposed which takes the distance between the subject and the 3D mesh into consideration. The adaptive threshold is calculated using the formula below:

\begin{equation}
thres = \frac{{h*D_k^{ij}}}{{||{I_k^{ij}} - {I_{k - 1}^{ij}}||_2}}
\end{equation}
where $h$ is a parameter controlling the range of the threshold, and $D_k^{ij}$ represents the current distance of the subject to the 3D mesh.  Note that the calculated velocity threshold will be higher if the subject is far away from the 3D mesh.

Using the adaptive threshold calculated above, all the intersection points can be categorized into fixation points or saccade points. However, these fixation points need to be clustered since fixations occur within a certain spatial interval presents a particular area of interest (AOI). The more fixations inside the AOI, the more attractive the AOI is \cite{10.1145/1518701.1518705}. By analyzing the clustered fixation points, subject's visual behaviour can be better explored. Most fixation clustering algorithms simply adopt the centroid of the fixations as the representation of the AOI, however, this is inappropriate due to the existence of noise points, which might reduce the accuracy of the result and influence the further investigation. 

To overcome this drawback, we apply the random walk mechanism to help identifying the center of the AOI since it can take the relationships between fixation points into consideration. Following the method proposed by Chen \textit{et al.} \cite{ChenExploringxiu}, we first calculate the transition probability of fixation points in a cluster using Euclidean distance between two fixation points and then incorporate the fixation density to evaluate the importance of a fixation point to the cluster. Later, the obtained transition probabilities and fixation densities are applied to the random walk mechanism to iteratively update the coefficients of each fixation until the center of each cluster is obtained. Finally, the center of each cluster will replace all the fixation points in the cluster for further investigations on subject's visual behaviour. Examples of  fixation classification and clustering results on 3D meshes are shown in Fig. \ref{fig5}.

\section{Data Analysis}
In this section, we first analyze the statistical characteristics of the database and then provide further investigations of the subject's viewing behaviour towards 3D meshes. During the analysis, the 32 3D meshes are divided into four different classes according to their types \cite{doi:10.1111}: Humans, Animals, Mechanical Parts and Familiar Objects. Each class contains 8 3D meshes. 

\subsection{Statistical Analysis}
Using the HTC Vive headset with aGlass eye-tracking module, we can collect the subject's 6DoF data and eye-movement data simultaneously while they are observing all 32 3D meshes. Using the data processing method mentioned above, the fixation points of all subjects on each 3D mesh are obtained. The average number of fixation points collected from the same subject on all 32 3D meshes is shown in Fig. \ref{fig15} (a). In the figure, the $X$ axis represents the 21 subjects and the $Y$ axis denotes the average number of fixation points.  According to Fig. \ref{fig15} (a), the average numbers of fixation points collected from the same subject on all 3D meshes are mainly between 20-28.

Besides analyzing the average number of fixation points for each subject, we also investigate subjects' moving directions when they start to observe the 3D mesh from starting points.
The result of the investigation is shown in Fig. \ref{fig15} (b). It can be observed from the figure that most of the subjects tend to walk in their left direction when they start to observe the 3D mesh and this phenomenon is more obvious for subjects at starting point $S_1$ and $S_3$. This indicates that subjects might have a left preference when observing 3D meshes, which might be an important cue for work related to head movement predicting in 6DoF environment.

Since subjects' heads and bodies are allowed to move freely while they are observing 3D meshes, the distribution of the head position is very important for exploring their visual attention behaviour. The corresponding head positions of the collected fixation points for all 3D meshes are obtained and the distribution of the head positions is shown in Fig. \ref{fig15} (c). Note that the 3D mesh is presented at the center of the virtual scene which coordinate is set to (0,1.5,0) and the $Y$ axis represents the height of the subject's head above the ground. According to the figure, it can be observed that the subjects' head positions are distributed evenly around the center, indicating that the proposed database collects the 6DoF data and eye-movement data for 3D meshes from $360^{\circ}$ viewing directions.

\subsection{Inter-Observer Variation}
Besides statistical analysis about the proposed database, we also investigate the inter-observer variation. Wang \textit{et al.} \cite{3272127} provided evidence indicating that subjects tend to generate similar fixation patterns for the same shape than for different ones. To assess whether this phenomenon exists in the proposed database, we conduct analysis in a similar way to Wang \textit{et al.} \cite{3272127}. 

Before the analysis, we generate the fixation density map for each subject on all 3D meshes. Following the fixation density map generation process introduced in \cite{doi:10.1111}, a Gaussian distribution is projected on the mesh with the standard deviation of the Gaussian distribution set to 35, corresponding to approximately $1^{\circ}$ of visual angle in our experiment. The fixation density map $M_i^j$ is obtained by summing the contributions of all fixation points collected from subject $i$ on mesh $j$. 

During the analysis, Pearson's Linear Correlation Coefficient (PLCC) is applied to measure the similarity between two different fixation density maps as in \cite{doi:10.1111}. The similarity value $S_{i{}_1{i_2}}^{{j_1}{j_1}}$ between fixation density map $M_{i_1}^{j_1}$ and $M_{i_2}^{j_1}$ of two subjects $i_1$ and $i_2$ for the same mesh $j_1$ is calculated first, which represents the agreement between subjects on the same mesh. Then the similarity value $S_{i{}_1{i_2}}^{{j_1}{j_2}}$ is computed between fixation density map $M_{i_1}^{j_1}$ and $M_{i_2}^{j_2}$ which denotes the agreement between subjects across different meshes. Note that $j_1 \ne j_2$ and $i_1 \ne i_2$. If $S_{i{}_1{i_2}}^{{j_1}{j_1}} > S_{i{}_1{i_2}}^{{j_1}{j_2}}$ for most pairs of subjects, we can come to the conclusion that similar fixation pattern is more likely to happen when subjects are observing the same mesh compared with observing different ones. 

\begin{figure}[!t]
	\begin{center}
		\includegraphics[width=8.7cm]{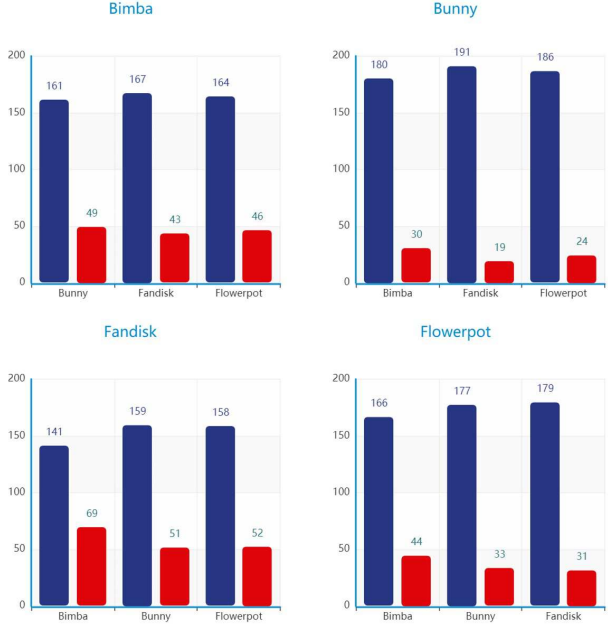}	
	\end{center}
	\caption{Inter-observer variation results. The blue bars represent the number of situations that $S_{i{}_1{i_2}}^{{j_1}{j_1}}$ is larger than $S_{i{}_1{i_2}}^{{j_1}{j_2}}$ while the red bars denote the opposite. The $p$-value for the rejection of the null hypothesis is also provided.}
	\label{fig6}
\end{figure}

In our experiment, we randomly select one mesh from each class for inter-observer variation calculation. For each combination of the 4 selected meshes (Bimba, Bunny, Fandisk, and Flowerpot), we calculate the corresponding  $S_{i{}_1{i_2}}^{{j_1}{j_1}}$ and $S_{i{}_1{i_2}}^{{j_1}{j_2}}$ for each possible pair of all 21 subjects participated in the experiment. Then we run the $t$-test to examine whether the difference between $S_{i{}_1{i_2}}^{{j_1}{j_1}}$ and $S_{i{}_1{i_2}}^{{j_1}{j_2}}$ is significant. The null hypothesis is that the mean similarity score $S_{i{}_1{i_2}}^{{j_1}{j_1}}$ is equal to the mean similarity score $S_{i{}_1{i_2}}^{{j_1}{j_2}}$.

The results of the analysis are presented in Fig. \ref{fig6}. The $p$-values presented below  also reject the null hypothesis, indicating that subjects tend to generate similar fixation patterns for the same mesh than for different ones. This investigation also confirms what has been observed in \cite{doi:10.1111, 3272127}.

\subsection{Attention Bias}
Since subjects typically show remarkable viewing tendencies toward the visual stimuli, analyzing their attention bias is an important way for exploring their visual attention behaviour. 

Center bias presents the phenomenon that fixation points are more likely to concentrate on the center of the image and has been widely observed from many eye-tracking experiments for 2D images and videos \cite{MeurA}. It is an important feature for the HVS and has been applied to many image saliency detection algorithms to help improve the detection accuracy. To examine whether this phenomenon exists in 3D meshes, we use the collected data to help the investigation. 

\begin{figure}[!t]
	\begin{center}
		\includegraphics[width=9.2cm]{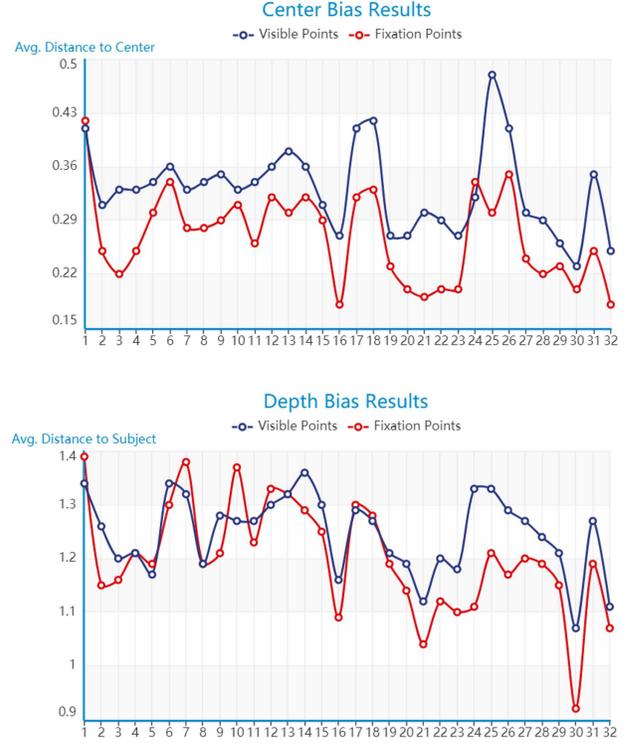}	
	\end{center}
	\caption{Center bias results and Depth bias results for 3D meshes.The top figure represents the results for analyzing the center bias while the bottom figure denotes the results for depth bias analysis. The red line represents the average distance of the fixation points towards the center/subject's head position for each mesh while the blue line denotes the average distance of all visible points towards the center/subject's head position for each mesh.}
	\label{fig8}
\end{figure}

In order to analyze the center bias, we calculate the visible points $V^{PO}$ on the mesh using the 6DoF data $(P, O)$ and obtain the corresponding fixation points $F^{PO}$. Note that $F^{PO}$ contains several fixation points collected from all subjects using the 6DoF data $(P, O)$. Then we calculate the center of the visible points $V^{PO}$ as $V_m^{PO}$. To quantitatively evaluate the degree of center bias for the mesh, we measure the average distance $D_F^{PO}$ between fixation points $F^{PO}$ and the center of the visible points $V_{m}^{PO}$ using the formula below:

\begin{equation}
\label{Eq:13}
{D_F^{PO}} = \sum\nolimits_{t = 1}^T {\frac{{||F_t^{PO} - V_m^{PO}||_2}}{T}}  
\end{equation}
where $F_t^{PO}$ represents the $t$-th fixation point in $F^{PO}$ and $T$ denotes the total number of fixation points in $F^{PO}$.

Besides calculating $D_F^{PO}$, the average distance $D_V^{PO}$ between the visible points $V^{PO}$ and the center $V_{m}^{PO}$ can also be calculated using Eqn. \ref{Eq:13}. The results of the calculated $D_F^{PO}$ and $D_V^{PO}$ for each 3D mesh is shown in the top of Fig. \ref{fig8} with the red line representing $D_F^{PO}$ and the blue line denoting $D_V^{PO}$. According to the figure, it can be observed that $D_F^{PO}$ is smaller than $D_V^{PO}$ for most of the 3D meshes, indicating that the fixation points are more concentrated to the center. This investigation demonstrates the existence of center bias preference for 3D meshes.  

Apart from the center bias preference, we also analyze whether subjects are more likely to be attracted by points that are close to them, which is denoted as depth bias. Depth bias is a commonly used hypothesis for many RGBD saliency detection frameworks. It suggests that objects near the subjects are more likely to be noticed. Following the method mentioned above, by changing the center of visible points $V_m^{PO}$ into the subject's head position, we can use Eqn. \ref{Eq:13} to investigate whether the depth bias exists for 3D meshes. The average distance between the visible points $V^{PO}$ and the subject's head position (blue line) as well as the average distance between the fixation points $F^{PO}$ and the subject's head position (red line) for all meshes are shown at the bottom of Fig. \ref{fig8}. According to the figure, for 22 out of 32 3D meshes, the average distance between the fixation points $F^{PO}$ and the subject's head position is lower than the average distance between the visible points  $V^{PO}$ and the subject's head position, but the depth bias preference is not as obvious as the center bias preference. 

\begin{figure}[!t]
	\begin{center}
		\includegraphics[width=9.2cm]{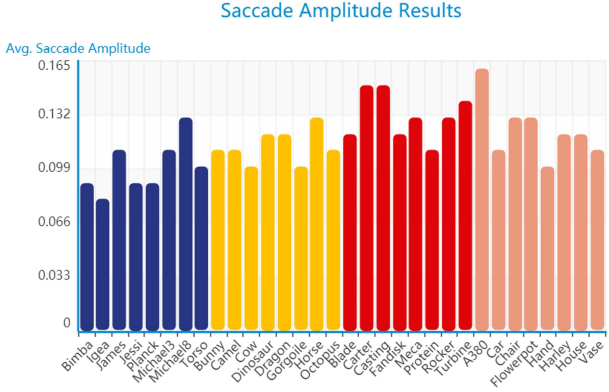}	
	\end{center}
	\caption{Average saccade amplitude for each 3D mesh.The blue bars, yellow bars, red bars and pink bars represent the average saccade amplitudes for Human class, Animal class, Mechanical class and Familiar Object class respectively.}
	\label{fig9}
\end{figure}

Besides investigating the bias and depth bias preference, the characteristics of saccades also need to be analyzed. Saccade represents the shift from one fixation point to another and plays an important role in human visual behaviour. Saccade amplitude is an important property of saccade and it represents the angular distance between two consecutive fixation points. To analyze the saccade amplitude in the proposed database, we calculate the average saccade amplitude of all subjects on each 3D mesh and the results are shown in Fig. \ref{fig9}. Note that the results are grouped by 3D meshes' class.  According to Fig. \ref{fig9}, it can be observed that the average saccade amplitude for all meshes belonging to Human class (blue bars) remains the lowest, while the average saccade amplitude for all meshes in Mechanical class (red bars) is the highest. The average saccade amplitude for Animal class (yellow bars) and Familiar Object class (pink bars) are slightly higher than that of the Human class but lower than the Mechanical class. This might indicate that when observing Human and Animal meshes, subjects are more likely to be attracted by small regions on the mesh and focus their attention on them, making the average saccade amplitude relatively lower. For Mechanical class, since it is unfamiliar to subjects, subjects are more likely to observe the whole mesh instead of focusing on a small region, leading to higher average saccade amplitude. 

\subsection{Viewing Direction Dependence}
The subjects' movements will change their viewing directions toward the 3D mesh.  As they move, some parts of the mesh become visible to subjects while the other parts become invisible. In this section, we analyze whether the subjects' viewing directions will have an influence on their visual attention distribution. 

To investigate the viewing direction dependence, Wang \textit{et al.} \cite{3272127} have conducted a similar analysis  which compares the difference between two fixation density maps collected from different viewing directions. According to their experiment, subjects are asked to fix their heads at a certain position and only seven fixed viewing directions are provided. Their experimental results suggest that most of the 3D meshes have significant viewing direction dependence. Besides Wang \textit{et al.} \cite{3272127}, Lavoue \textit{et al.} \cite{doi:10.1111} also find that different movements of the camera may produce very different fixation density maps. Considering the studies mentioned above, we wondered whether the viewing direction dependence exists in our database which allows the subjects to freely move their heads during 3D mesh observation.

\begin{table}[!t]
	\caption{Viewing direction dependence results.}
	\centering
	\scalebox{0.85}{
		\begin{tabular}{cccccc}
			\hline
			Category                                                                                         & Name     & CC                        & Category                                                                                            & Name      & CC   \\ \hline  \hline
			\multicolumn{1}{c|}{\multirow{8}{*}{Humans}}                                                     & Bimba    & \multicolumn{1}{c|}{0.26} & \multicolumn{1}{c|}{\multirow{8}{*}{Animals}}                                                       & Bunny     & 0.32 \\
			\multicolumn{1}{c|}{}                                                                            & Igea     & \multicolumn{1}{c|}{0.46} & \multicolumn{1}{c|}{}                                                                               & Camel     & 0.48 \\
			\multicolumn{1}{c|}{}                                                                            & James    & \multicolumn{1}{c|}{0.32} & \multicolumn{1}{c|}{}                                                                               & Cow       & 0.44 \\
			\multicolumn{1}{c|}{}                                                                            & Jessi    & \multicolumn{1}{c|}{0.52} & \multicolumn{1}{c|}{}                                                                               & Dinosaur  & 0.33 \\
			\multicolumn{1}{c|}{}                                                                            & Planck   & \multicolumn{1}{c|}{0.28} & \multicolumn{1}{c|}{}                                                                               & Dragon    & 0.52 \\
			\multicolumn{1}{c|}{}                                                                            & Michael3 & \multicolumn{1}{c|}{0.47} & \multicolumn{1}{c|}{}                                                                               & Gorgoile  & 0.49 \\
			\multicolumn{1}{c|}{}                                                                            & Michael8 & \multicolumn{1}{c|}{0.34} & \multicolumn{1}{c|}{}                                                                               & Horse     & 0.60 \\
			\multicolumn{1}{c|}{}                                                                            & Torso    & \multicolumn{1}{c|}{0.45} & \multicolumn{1}{c|}{}                                                                               & Octopus   & 0.47 \\ \hline
			\multicolumn{1}{c|}{\multirow{8}{*}{\begin{tabular}[c]{@{}c@{}}Mechanical\\ Parts\end{tabular}}} & Blade    & \multicolumn{1}{c|}{0.45} & \multicolumn{1}{c|}{\multirow{8}{*}{\begin{tabular}[c]{@{}c@{}}Familiar\\ Objects\end{tabular}}} & A380      & 0.51 \\  
			\multicolumn{1}{c|}{}                                                                            & Carter   & \multicolumn{1}{c|}{0.54} & \multicolumn{1}{c|}{}                                                                               & Car       & 0.50 \\
			\multicolumn{1}{c|}{}                                                                            & Casting  & \multicolumn{1}{c|}{0.30} & \multicolumn{1}{c|}{}                                                                               & Chair     & 0.49 \\
			\multicolumn{1}{c|}{}                                                                            & Fandisk  & \multicolumn{1}{c|}{0.29} & \multicolumn{1}{c|}{}                                                                               & Flowerpot & 0.69 \\
			\multicolumn{1}{c|}{}                                                                            & Meca     & \multicolumn{1}{c|}{0.49} & \multicolumn{1}{c|}{}                                                                               & Hand      & 0.46 \\
			\multicolumn{1}{c|}{}                                                                            & Protein  & \multicolumn{1}{c|}{0.54} & \multicolumn{1}{c|}{}                                                                               & Harley    & 0.38 \\
			\multicolumn{1}{c|}{}                                                                            & Rocker   & \multicolumn{1}{c|}{0.66} & \multicolumn{1}{c|}{}                                                                               & House     & 0.52 \\
			\multicolumn{1}{c|}{}                                                                            & Turbine  & \multicolumn{1}{c|}{0.55} & \multicolumn{1}{c|}{}                                                                               & Vase      & 0.46 \\ \hline
	\end{tabular}}
	\label{table3}
\end{table}

\begin{figure*}[!t]
	\begin{center}
		\includegraphics[width=17.0cm]{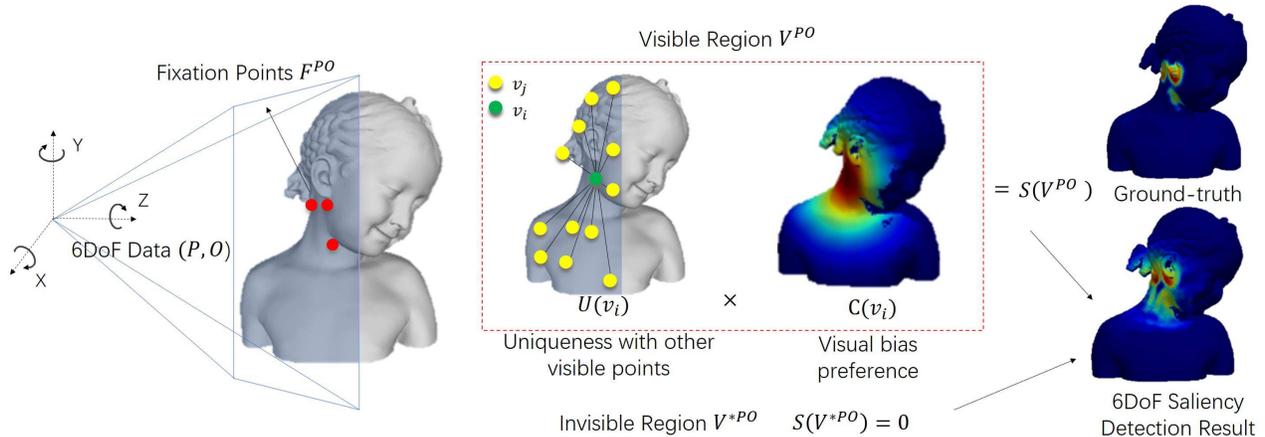}	
	\end{center}
	\caption{Framework of the proposed 6DoF mesh saliency detection algorithm. In the left part, subject's 6DoF data $(P, O)$ decides the visible points $V^{PO}$ (blue shading) and the invisible points $V^{*PO}$of the mesh. The red points are the collected fixation points $F^{PO}$ which generate the ground-truth map shown in the upper right. The red rectangle presents the process of saliency calculation for the visible points $V^{PO}$. The green point represents the visible point $v_i$ in $V^{PO}$ while the yellow points denote the other visible points $v_j$ in $V^{PO}$.}
	\label{fig13}
\end{figure*}

Different from the experiment proposed by Wang \textit{et al.} \cite{3272127} and Lavoue \textit{et al.} \cite{doi:10.1111}, for each 3D mesh, the collected eye-movement data in our experiment covers approximately  $360^{\circ}$ viewing directions. Since different viewing directions lead to very different visible regions, comparing fixation density maps that are collected from very different viewing directions is meaningless. In this experiment, we only consider a subset of the database in which the maximum difference of any two viewing directions is limited to $90^{\circ}$. The height of the head position in the subset is limited to 1.6m since the majority of data was collected under this height.

Since there are many data in the subset, for a specific 3D mesh, we randomly select 10 viewing directions in the subset with the corresponding fixation density maps. Then we compute the similarity of the selected fixation density map (FDM) in pairs using correlation coefficient value and record their difference in viewing directions. Later, we estimate the relationship between FDM similarity and their difference in viewing directions using correlation coefficient to examine whether the viewing dependence exists. The correlation coefficient calculated for each 3D mesh is shown in Table. \ref{table3}. Note that for each 3D mesh, the calculation has been repeated 100 times to ensure the stability of the results. 

According to Table. \ref{table3}, it can be observed that for 23 out of 32 3D meshes, there is a significant correlation between the FDM similarity and the difference in viewing directions (with correlation coefficient larger than 0.4), which indicates that with the increasing difference in viewing directions, the similarity of the fixation density maps decreases. Besides, we can also learn from the results that most of the Human mesh have lower viewing direction dependence, such as the Bimba mesh, suggesting that subjects might be attracted by similar regions on Human meshes even the viewing direction changes. This also verifies the investigations provided in \cite{3272127}.

\section{6DoF mesh saliency detection}
In this section, we provide a 6DoF mesh saliency detection algorithm based on the aforementioned analysis and a framework of the proposed algorithm is shown in Fig.\ref{fig13}. Besides, an ablation study is conducted to illustrate the effectiveness of the uniqueness calculation and the bias preference.

\subsection{The proposed algorithm}

According to the analysis mentioned above,  we observed the existence of depth bias in 3D mesh observation and that point shown in the center of the FoV is more likely to attract subject's visual attention than other points. Moreover, we also investigated the influence of viewing direction towards the 3D mesh observation which suggests that the subject's head movement will have a great impact on mesh saliency detection. Based on these observations, we develop a computational model for detecting mesh saliency using the subject's 6DoF data. To the best of our knowledge, no algorithm has been proposed using this information, and we hope this `dynamic' mesh saliency detection model can be applied to `dynamic' applications such as dynamic mesh compression \cite{7405340} and interaction \cite {40774762}.

Traditional mesh saliency detection algorithms do not take the visibility of the mesh into consideration and assign saliency values to invisible points, which is impractical for real applications since the observation towards 3D meshes is usually a dynamic process. Besides, it has been illustrated in \cite{doi:10.1111} that the subject's fixations in the dynamic scene might not directly related to 3D geometry but to the changes in shadowing/reflection or  other factors that occur during the movements, so the subject's head movement will greatly influence the saliency detection results towards the 3D mesh.  

To address this problem, we need to take the visibility of a region on the 3D mesh into consideration. Whether a point on the mesh is visible or invisible is depending on the position of the mesh and the subject's 6DoF data (head position and orientation). For a 3D mesh and a 6DoF data $(P, O)$, the painter's algorithm \cite{10.5555/265140} is used to solve the visibility problem and obtain a binary field to distinguish which part of the mesh is visible. As is shown in the left of Fig. \ref{fig13}, the visible points $V^{PO}$ of the mesh is covered by blue shading.

With the obtained visible points $V^{PO}$ on the 3D mesh, we calculate the low-level feature Fast Point Feature Histogram (FPFH) \cite{5152473} for each visible point in $V^{PO}$ since it has been proven to work well for 3D saliency detection \cite{6751558}. With the obtained FPFH descriptor for each visible point, we calculate the dissimilarity between two visible points $v_i$ and $v_j$. Unlike other algorithms which use the Chi-Squared distance to calculate the dissimilarity, in this paper, the Bhattacharyya distance is utilized since it avoids the possible situation that the denominator of the Chi-Squared method might be equal to zero. The dissimilarity between $v_i$ and $v_j$ can be calculated using the formula below:

\begin{equation}
Dis({v_i},{v_j}) = -\log(\sum\limits_{n = 1}^N {\sqrt{f_{ni}f_{nj}}})
\end{equation}
where $Dis({v_i},{v_j})$ represents the dissimilarity value. $N$ is the number of bins in the FPFH descriptor which is set to 33. $f_{ni}$ and $f_{nj}$ represent the $n$-th bin of the FPFH descriptor for visible points $v_i$ and $v_j$ respectively.

Since it has been widely acknowledged that regions differ from their surroundings are more likely to attract visual attention, we calculate the uniqueness of the visible point $v_i$ in $V^{PO}$ using the formula below. Note that the uniqueness value of  invisible points belonging to $V^{*PO}$ will be set to 0. An intuitive presentation of the uniqueness calculation process is shown in the left of the red rectangle in Fig. \ref{fig13}. The green point denotes the visible point $v_i$ and the yellow points represent the other visible points on the mesh.

\begin{equation}
U({v_i}) = {1 - \exp ( - \frac{1}{V^{PO}}\sum\limits_{{v_j} \in V^{PO}} {\frac{{Dis({v_i},{v_j})}}{{1 + ||{v_i} - {v_j}||_2}}} )}  
\end{equation}
where $V^{PO}$ denotes the visible points on the mesh using 6DoF data $(P, O)$, $U({v_i})$ is the calculated uniqueness for visible point $v_i$.

\begin{figure}[!t]
	\centering
	\subfigure[]{\includegraphics[height=7.2cm]{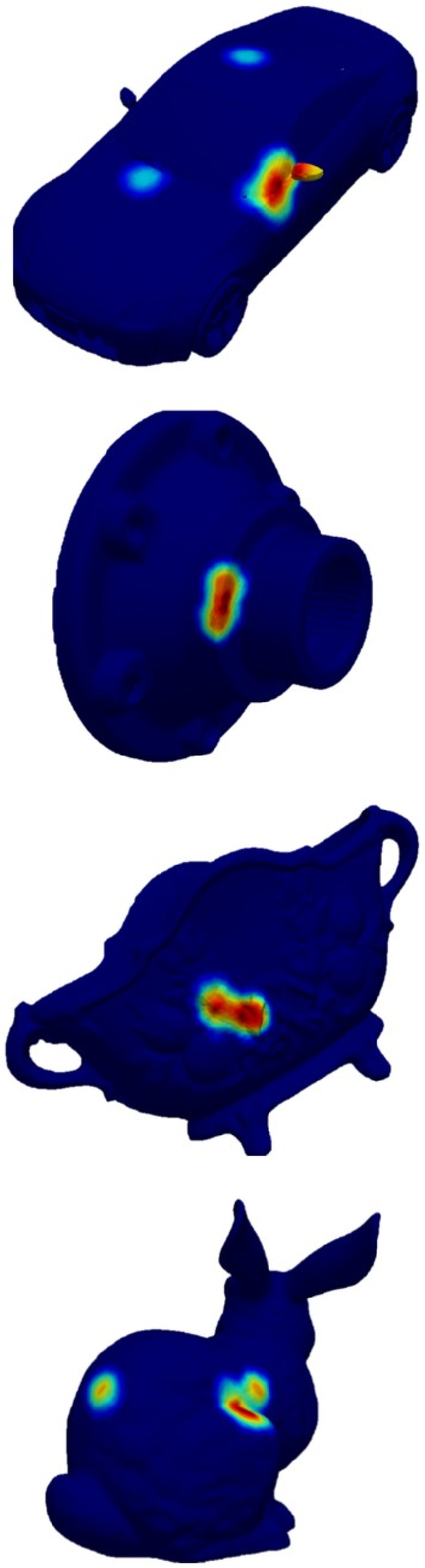}}
	\subfigure[]{\includegraphics[height=7.2cm]{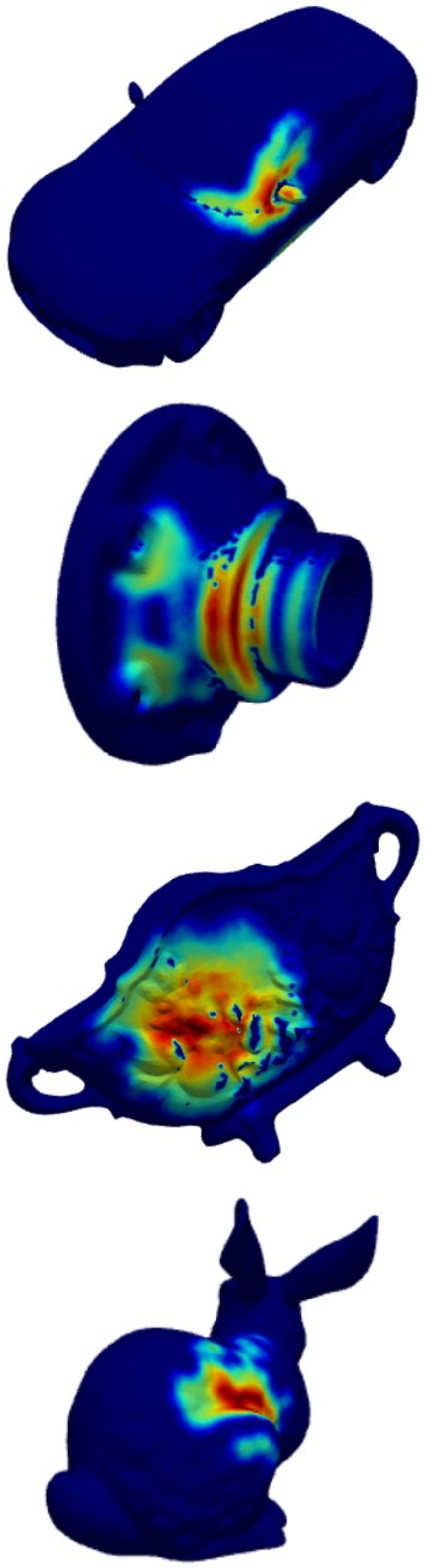}}
	\subfigure[]{\includegraphics[height=7.2cm]{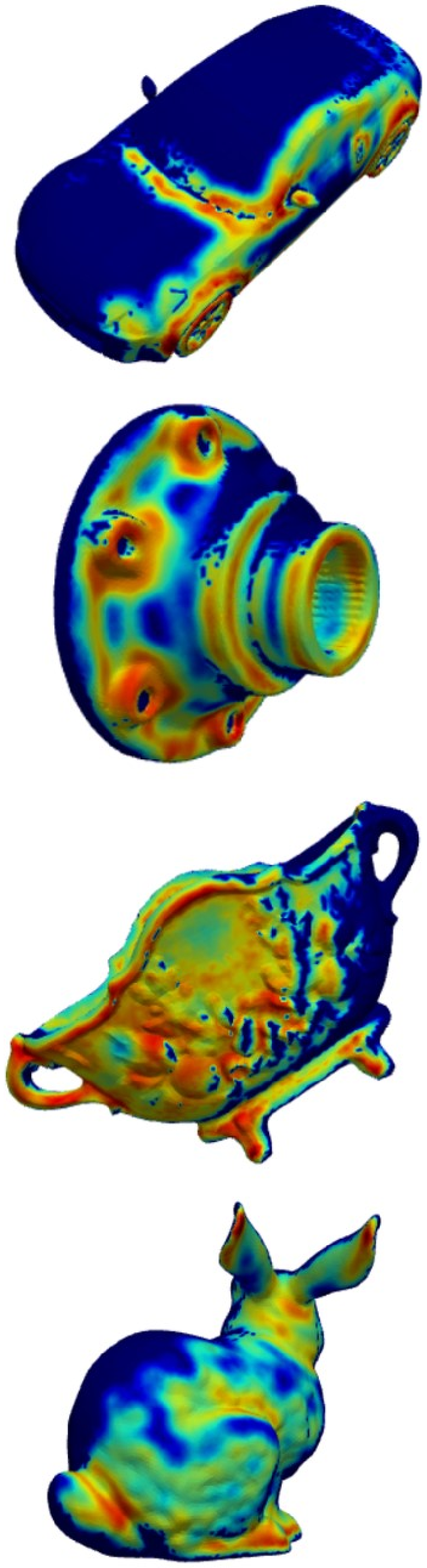}}
	\subfigure[]{\includegraphics[height=7.2cm]{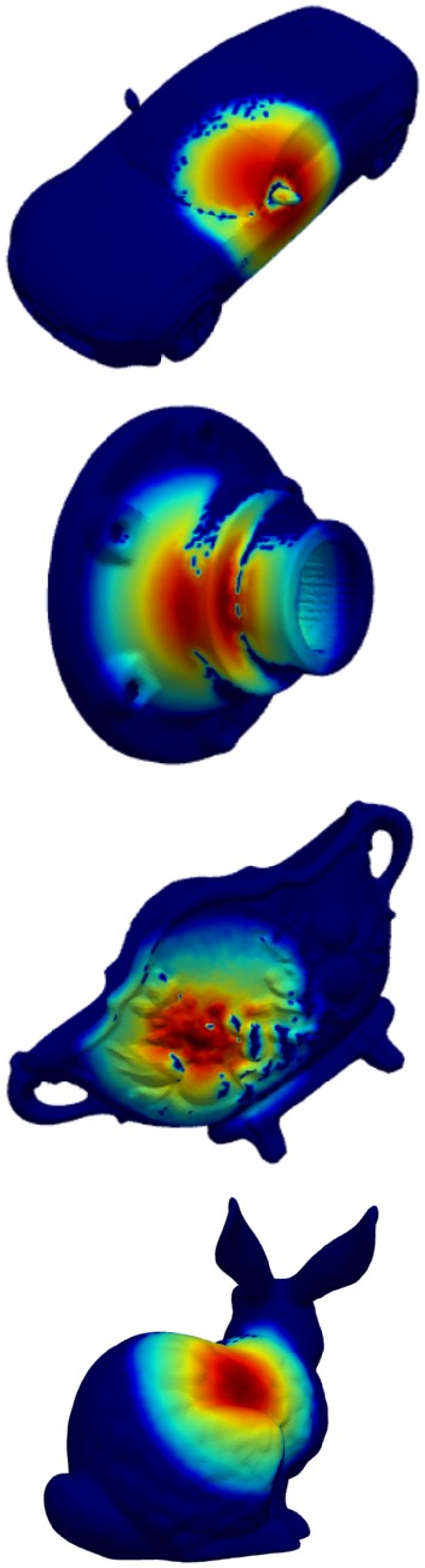}}
	\caption{Results of the 6DoF mesh saliency detection algorithm. Column (a) is the collected fixation density map  while (b) is the saliency detection results of the proposed 6DoF mesh saliency detection algorithm. Column (c) shows the results of using the proposed uniqueness calculation while (d) presents the results of simulating the bias preference. The red region suggests higher saliency value while the blue region indicates lower saliency value.}
	\label{fig10}
\end{figure}

As have been investigated in the analysis section, when subjects are moving around to freely observe 3D meshes, they have visual bias preference and is more likely to be attracted by points shown in the center of the FoV. Considering this impact, a refinement is needed to simulate this visual preference by taking the subject's 6DoF data $(P, O)$ into consideration. The refinement needs to encourage higher saliency values for points near the center of the FoV and decrease the saliency values for points which have larger distance to the center. The visual bias value of a visible point $v_i$ is calculated as the spatial distance between $v_i$ and the center of the visible points $V_m^{PO}$, which is defined using the formula below:
\begin{equation}
C({v_i}) = {\exp({-\frac{||{v_i} - {V_m^{PO}}||_2}{2\sigma^2}})} 
\end{equation}
where $C(v_i)$ represents the visual bias value for visible point $v_i$. Note that $\sigma$ is a controlling constant which is set to 0.2.

With the obtained uniqueness value $U(v_i)$ and visual bias value $C(v_i)$, the final saliency detection result $S(v_i)$ for visible point $v_i$ is calculated using the formula below:
\begin{equation}
S({v_i}) = U({v_i}) \times C(v_i)
\end{equation}

Using the formula above, we can predict the visual saliency map of a 3D mesh in any viewing direction and even predict the visual saliency map of a new 3D mesh. Examples of the 6DoF mesh saliency detection results using the proposed algorithm are shown in Fig. \ref{fig10}. According to the figure, it can be observed that the proposed algorithm captures the salient regions on the 3D mesh quite well but failed to detect salient regions that are far away from the center, this might be caused by the visual bias term in the formula that suppress the saliency value in these regions.

\section{6DoF mesh saliency evaluation}

\begin{figure*}[!t]
	\centering
	\subfigure[]{\includegraphics[height=4.3cm]{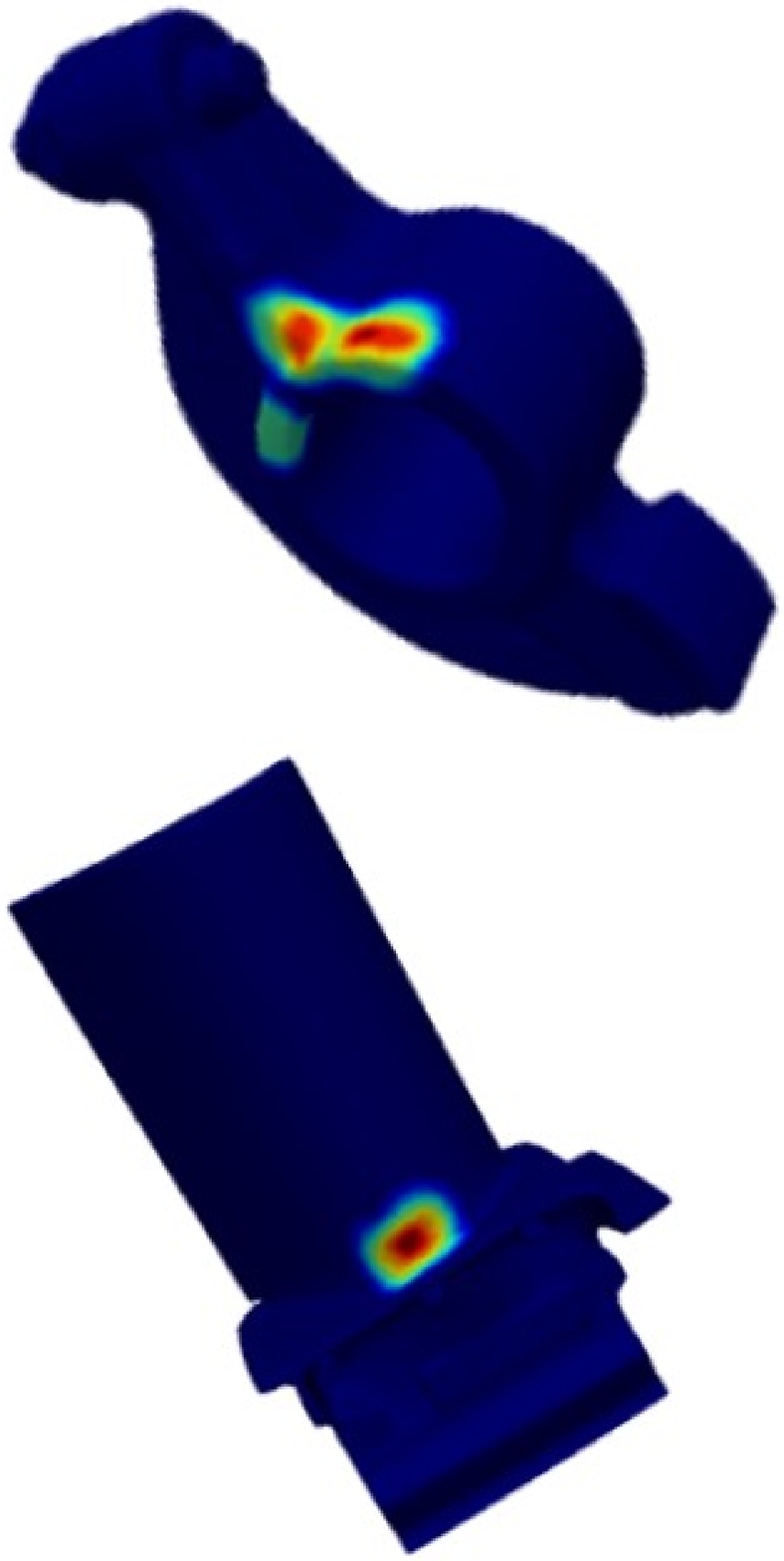}}
	\subfigure[]{\includegraphics[height=4.3cm]{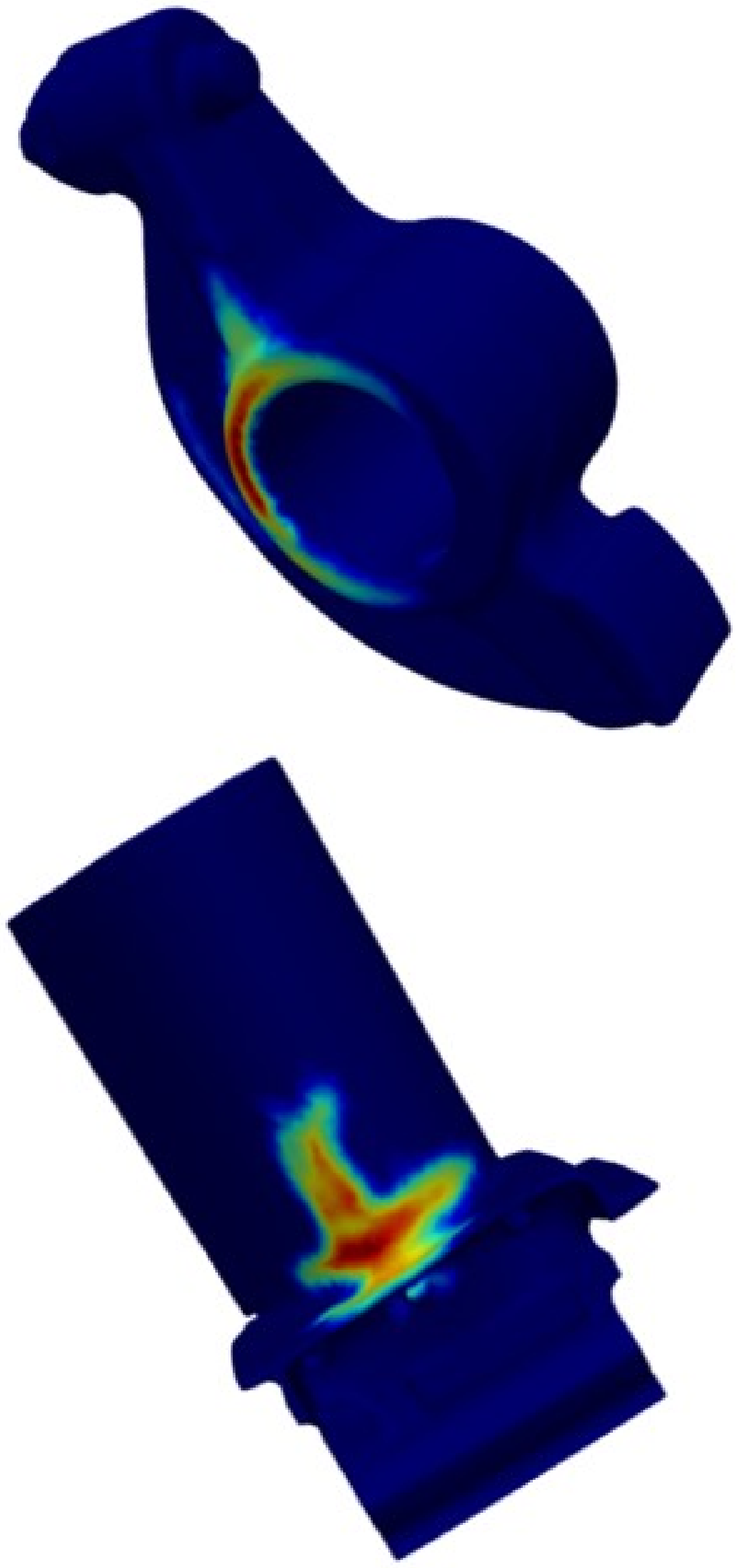}}
	\subfigure[]{\includegraphics[height=4.3cm]{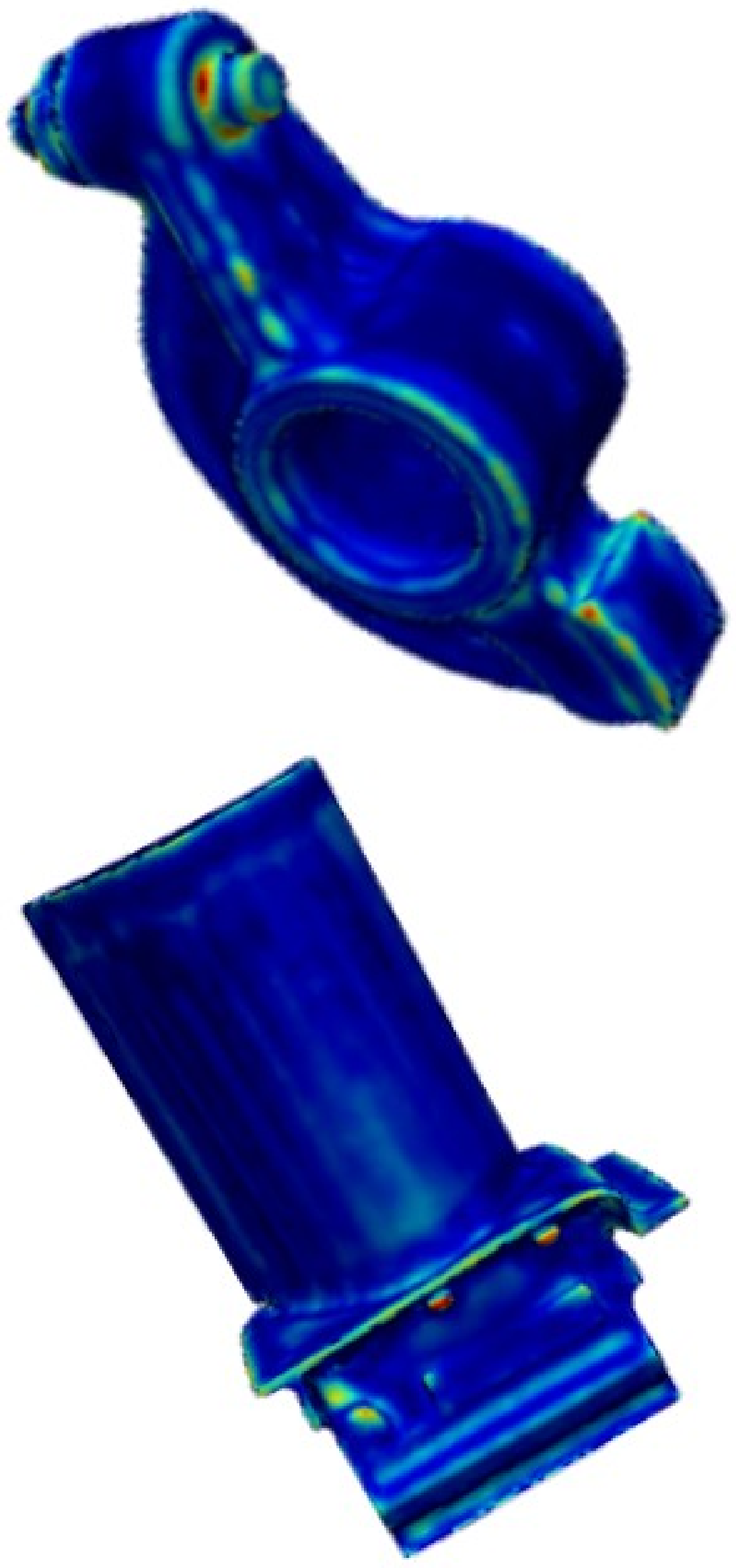}}
	\subfigure[]{\includegraphics[height=4.3cm]{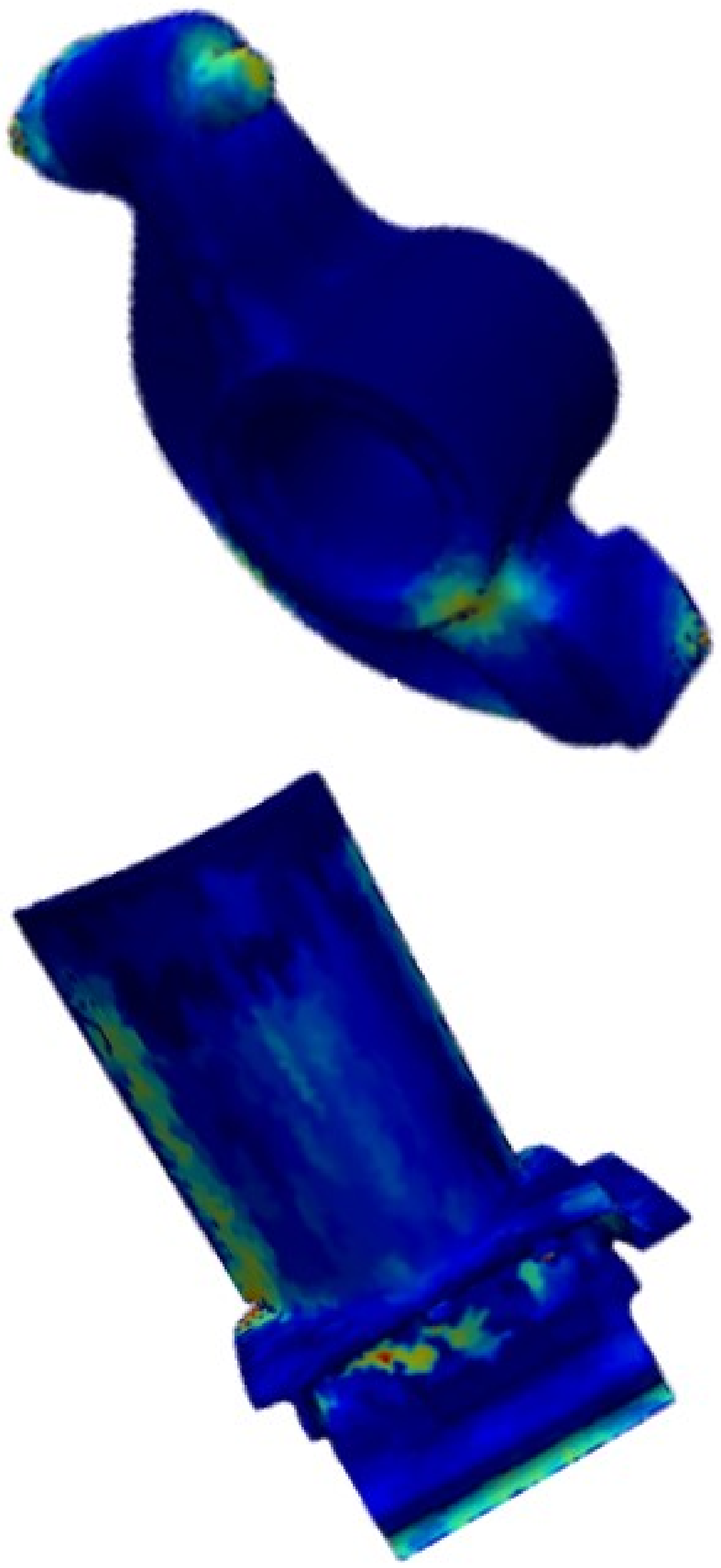}}
	\subfigure[]{\includegraphics[height=4.3cm]{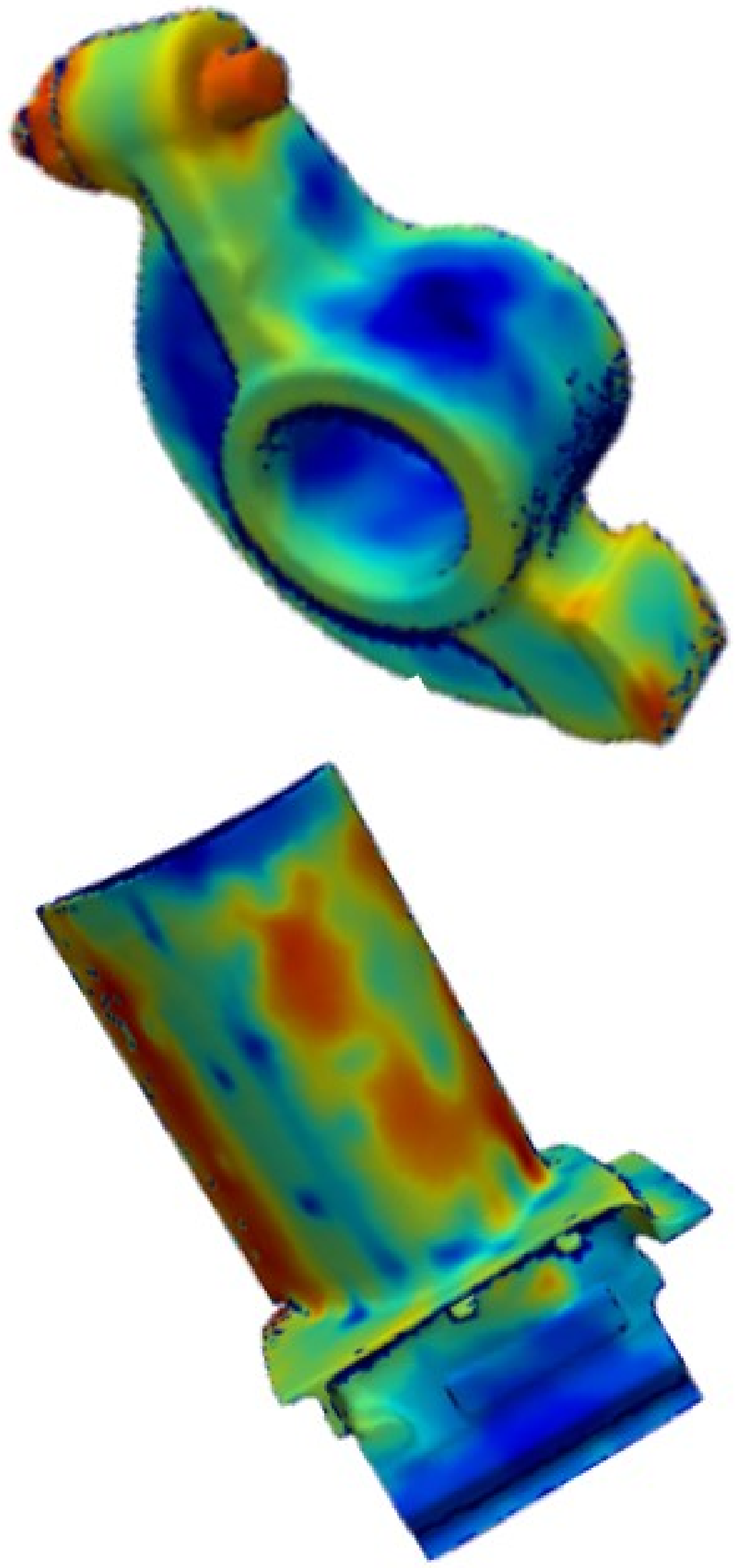}}
	\subfigure[]{\includegraphics[height=4.3cm]{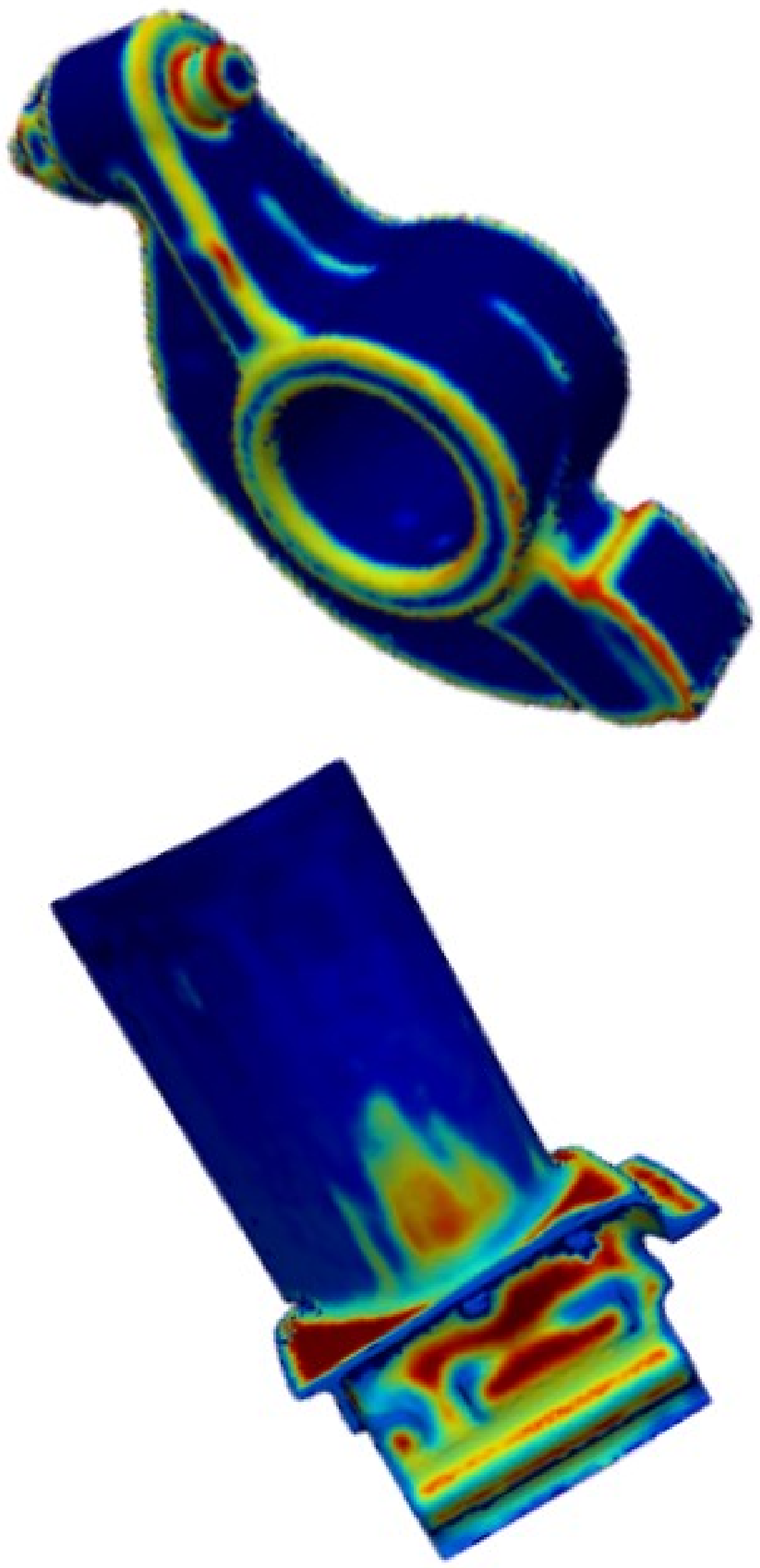}}
	\subfigure[]{\includegraphics[height=4.3cm]{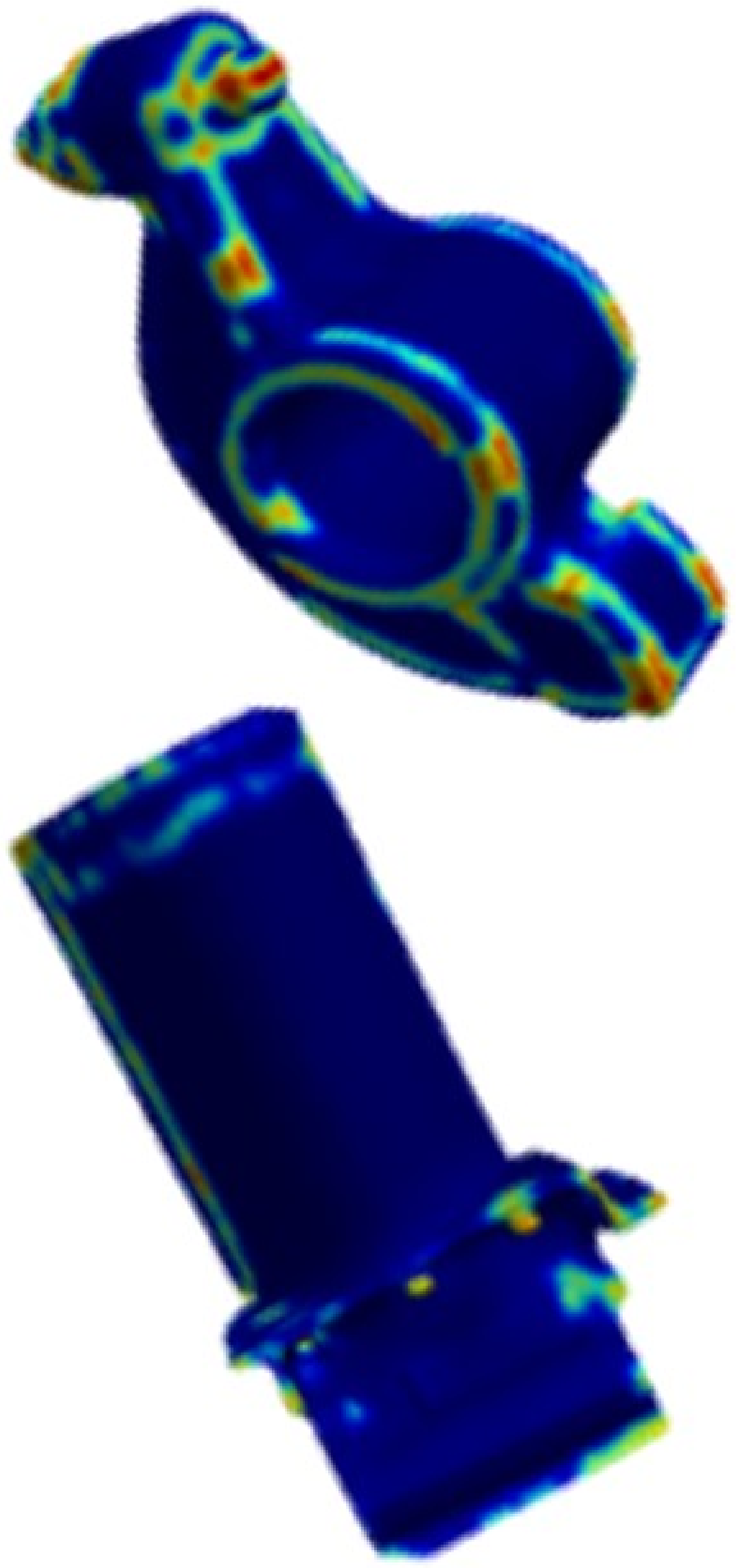}}
	\caption{Comparison of different 6DoF mesh saliency detection results. Column (a) and (b) are the ground-truth FDMs and the saliency detection results of the proposed 6DoF mesh saliency detection method. Column (c), (d), (e), (f) and (g) are the saliency detection results from Lee \textit{et al.} \cite{1073204.1073244}, Shtrom \textit{et al.} \cite{6751558}, Song \textit{et al.} \cite{song2014mesh}, Tasse \textit{et al.} \cite{7410384} and Ding \textit{et al.} \cite{8726371} respectively.}
	\label{fig11}
\end{figure*}

\subsection{Ablation study}
To better illustrate the effectiveness of the uniqueness calculation and the visual bias preference, an ablation study is conducted in which we report the 6DoF saliency detection results of using only uniqueness calculation and visual bias preference. The results of the ablation study are shown in Fig. \ref{fig10} (c) and (d).

\begin{figure}[!t]
	\begin{center}
		\includegraphics[width=8.0cm]{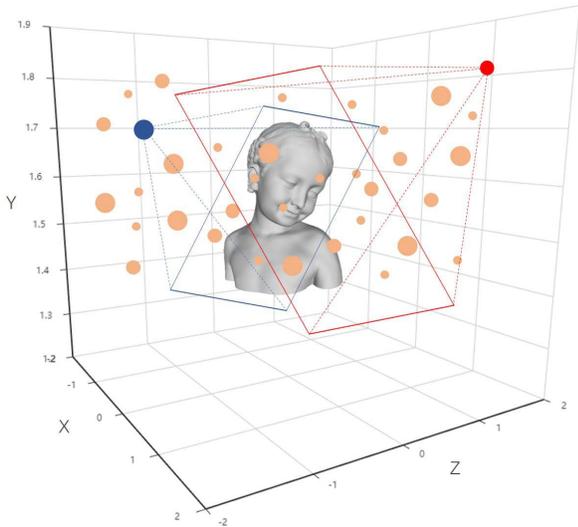}	
	\end{center}
	\caption{Distribution of the collected head positions for the mesh. The orange points represent the collected head positions with larger size indicating that more subjects have visited this head position during their observation. The red and blue points are examples of the collected head positions and the red/blue rectangle presents the subject's FoV.}
	\label{fig15}
\end{figure} 

According to the figure, the saliency detection results using uniqueness calculation highlight regions that are different from their surroundings but also assign saliency values to regions that near the edge of the FoV, leading to quite unsatisfying saliency detection results when compared with the ground-truth (Fig. \ref{fig10} (a)). The saliency detection results using visual bias preference assign higher saliency values to regions that near the center of the FoV but do not take the geometric feature of the mesh into consideration. By combing the uniqueness calculation with the visual bias preference, the saliency detection results in Fig. \ref{fig10} (b) are more similar to the ground-truth such as assigning higher saliency value to the rearview mirror in the car mesh, indicating better 6DoF mesh saliency detection performance.

Saliency evaluation metric helps to examine how well different saliency detection algorithms perform when compared to the collected human fixation data. Many saliency evaluation metrics have been proposed to help evaluating image saliency detection algorithms, such as the NSS metric which measures the mean value of the normalized saliency map at the fixation locations and the AUC metric which is based on calculating the true positive rate and false positive rate for the detection result. However, less saliency evaluation metric has been designed for evaluating 3D saliency detection algorithms and no evaluation metric has been proposed for 6DoF mesh saliency evaluation. In this paper, we focus on building a saliency evaluation metric for 6DoF mesh saliency detection. Since different subjects' movements may produce very different fixation density maps \cite{doi:10.1111}, to evaluate the performance of the 6DoF mesh saliency detection algorithm, we generate the ground-truth fixation density map for each mesh and each 6DoF data using the fixation points collected from all subjects. 

Previous evaluation metrics for 3D saliency detection use the Correlation Coefficient (CC) score and Saliency Error (SE) score \cite{8726371} to evaluate the performance of different 3D saliency detection algorithms. However, these metrics do not take the distribution of the saliency values on the 3D mesh into consideration. To address this problem, we also involve the Kullback-Leibler (KL) divergence to help evaluating the performance of 6DoF mesh saliency detection methods. Note that lower SE/KL score and higher CC score denote better saliency detection result.

During the evaluation, we first obtain the ground-truth fixation density map $G_w^j$ and the saliency detection result $R_w^j$ for the $w$-th 6DoF data collected on mesh $j$. For each collected head position, several subjects might have visited the same head position during their observation. Head position which has been visited by more subjects is more important than head position that has been visited by less subjects, and should be assignd higher weights during the evaluation. An example is shown in Fig. \ref{fig15}. Larger point indicates that more subjects have visited this head position during their observation. The overall evaluation result of the algorithm is computed as the weighted-sum of the saliency evaluation metric (CC, SE and KL), which is calculated using the formula below:

\begin{figure*}[!t]
	\begin{center}
		\subfigure[]{\includegraphics[width=16.6cm]{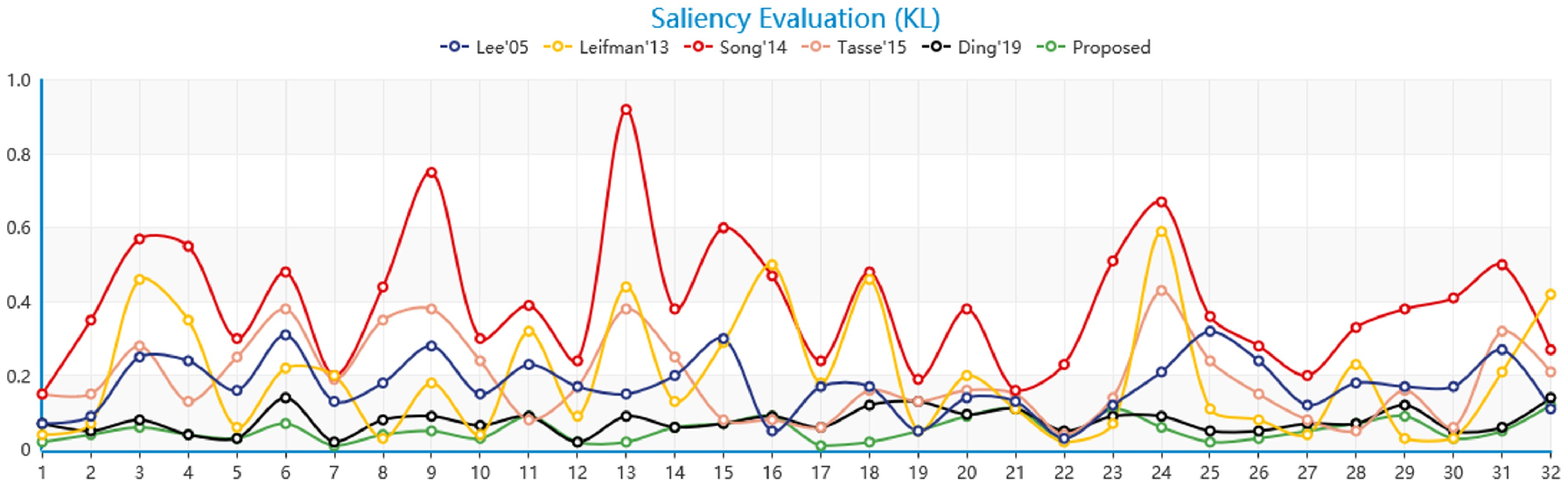}}	
		\subfigure[]{\includegraphics[width=16.6cm]{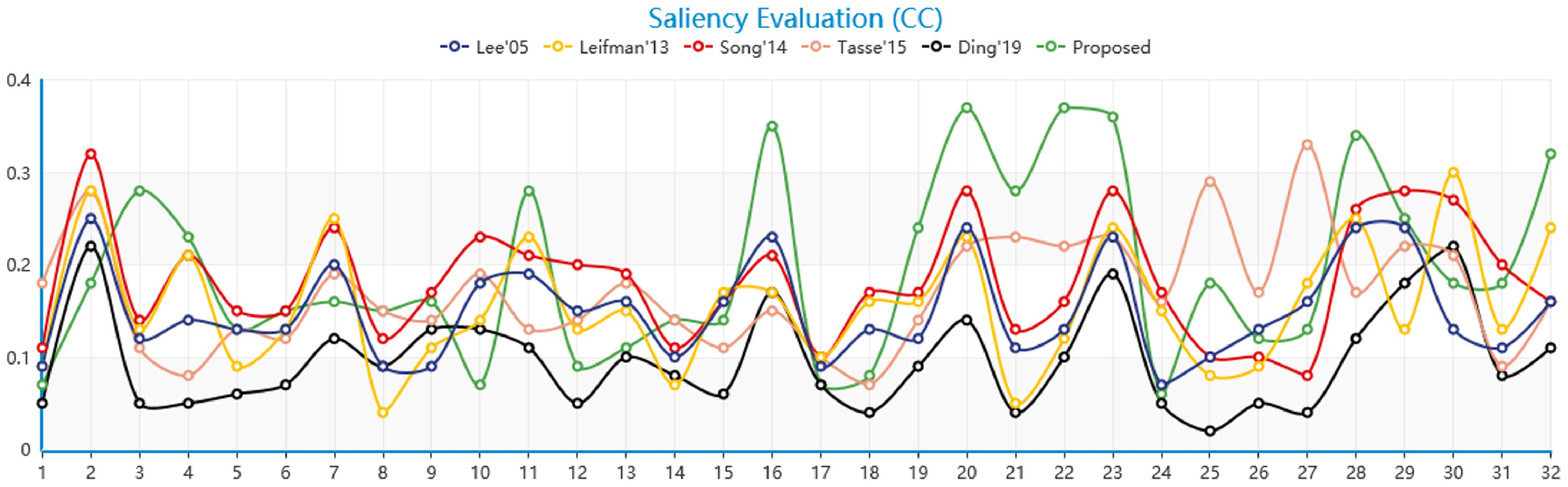}}	
		\subfigure[]{\includegraphics[width=16.6cm]{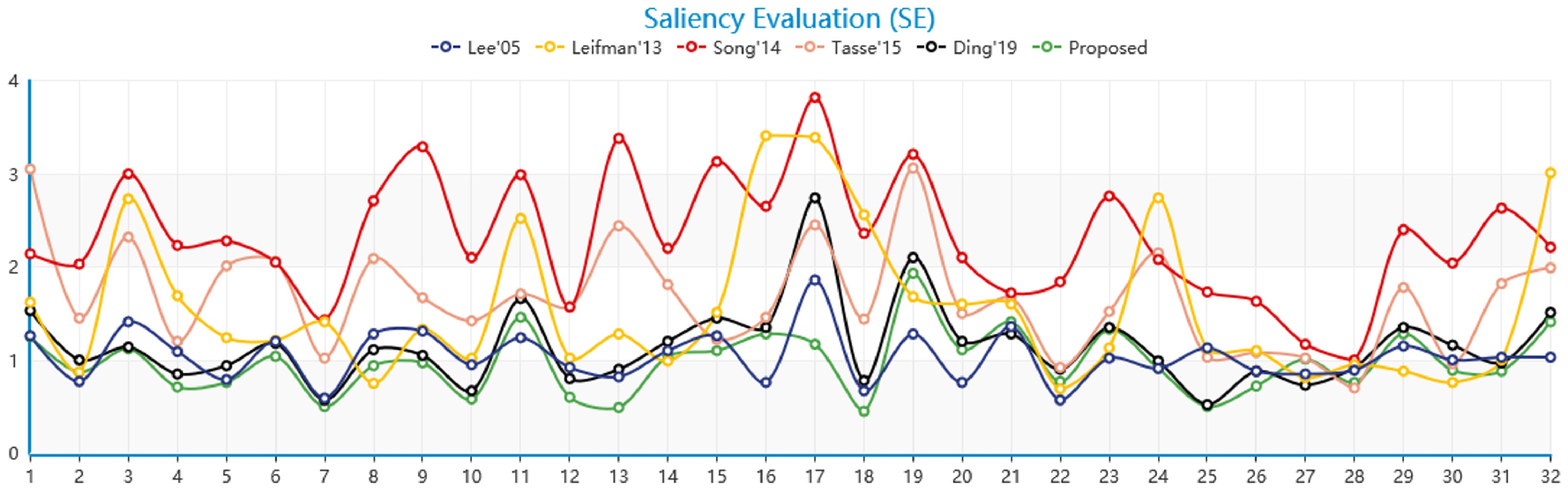}}		
	\end{center}
	\caption{Quantitative evaluation results for different saliency detection algorithms. (a), (b) and (c) represent the evaluation results using KL, CC and SE evaluation metrics. The green, blue, yellow, red, pink and black colours represent the evaluation results for the proposed 6DoF saliency detection algorithm, the algorithm proposed by Lee \textit{et al.} \cite{1073204.1073244}, Shtrom \textit{et al.} \cite{6751558}, Song \textit{et al.} \cite{song2014mesh}, Tasse \textit{et al.} \cite{7410384} and Ding \textit{et al.} \cite{8726371} respectively.}
	\label{fig12}
\end{figure*}

\begin{equation}
E_{Eval}^j = \frac{{\sum A_w*Eval(G_w^j, R_w^j)}}{{\sum A_w}}
\end{equation}
where $E_{Eval}^j$ represents the evaluation result for mesh $j$ using different evaluation metrics and $Eval \in \{KL, CC, SE\}$. $A_w$ represents the number of subjects that have visited the $w$-th head position during their observation.

With the obtained 6DoF saliency evaluation metric, we can evaluate the performance of the proposed 6DoF mesh saliency detection algorithm. However, due to the absence of 6DoF mesh saliency detection algorithms, several state-of-the-art 3D saliency detection methods are extended to help make comparisons with the proposed algorithm. These 3D saliency detection methods are summarized below:

\noindent \textbf{Lee \textit{et al.} \cite{1073204.1073244}}: Since curvature is an invariant property of the mesh, the curvature change on the mesh is considered in detecting mesh saliency. The algorithm first uses the Gaussian-weighted mean curvature with center-surround mechanism to calculate saliency values for the 3D mesh at multiple scales, and then uses a non-linear suppression operator which helps to combine saliency maps at different scales to generate the final saliency detection result.

\noindent \textbf{Shtrom \textit{et al.} \cite{6751558}}: The algorithm detects saliency values for the mesh hierarchically and for each level, a different neighbourhood is considered. In the algorithm, the low-level distinctness is first calculated to highlight delicate features, then the point association is applied which considers the regions near the foci of attention to be more interesting. Later, a high-level distinctness is computed for detecting large features and the saliency detection result is obtained by integrating the above three components using linear combination.

\noindent \textbf{Song \textit{et al. }\cite{song2014mesh}}: The algorithm detects the saliency values for the mesh using its spectral attributes. In this approach, the properties of the log-Laplacian spectrum of the mesh are first obtained and the irregularity of the spectrum is considered to be highly related to the saliency value of the mesh. Then the irregularity of the spectrum is captured and transferred to spatial domain in a multi-scale manner to obtain a more reliable saliency detection result.

\noindent \textbf{Tasse \textit{et al.} \cite{7410384}}: The algorithm is a cluster-based approach for point cloud saliency detection. Different from other approaches, the algorithm first decomposes the point cloud data into several clusters using fuzzy clustering algorithm and then computes the saliency value for each cluster using its uniqueness and spatial distribution. Later, the probabilities of points belonging to each cluster are calculated and the cluster-based saliency values are propagated to each point to obtain the saliency detection result.

\noindent \textbf{Ding \textit{et al.} \cite{8726371}}: The algorithm detects point cloud saliency by integrating local and global features. In this approach, the local distinctness is calculated based on the difference with local surrounding points, and the global rarity is obtained by using the random walk ranking method to propagate the cluster-level global rarity into each point. Later, an optimization framework is used to integrate the local and global cues to obtain the final saliency detection result.

\begin{table}[]
	\caption{Average saliency evaluation results}
	\begin{center}
		\scalebox{1.0}{
			\begin{tabular}{c|cccccccccccccccc}
				\hline
				Metric &Avg. KL    &Avg. CC    &Avg. SE     \\ \hline  \hline
				Lee \textit{et al.} \cite{1073204.1073244}  & 0.17 & 0.15    & 1.04    \\
				Shtrom \textit{et al.} \cite{6751558}  & 0.19    & 0.16    & 1.55     \\
				Song \textit{et al.} \cite{song2014mesh}  & 0.40    & 0.18    & 2.31   \\
				Tasse \textit{et al.} \cite{7410384}  & 0.19    & 0.17    & 1.67   \\
				Ding \textit{et al.} \cite{8726371}  & 0.08    & 0.10    & 1.15    \\
				Proposed  & \textcolor{red}{0.05}    & \textcolor{red}{0.20} & \textcolor{red}{0.98}  \\ \hline
		\end{tabular} }
	\end{center}
	\label{table4}
\end{table}

Since these algorithms are developed without using 6DoF information, to make a fair comparison, we take the visibility of the mesh into consideration. For each collected 6DoF data, the visibility of the points is a binary field and these 3D saliency detection algorithms' results will multiply with this binary field during the evaluation. Examples of the saliency detection results obtained by multiplying the binary field with the five algorithms mentioned above are shown in Fig. \ref{fig11}. Besides displaying the 6DoF mesh saliency detection results, we also provide a quantitative evaluation among these algorithms using the proposed 6DoF saliency evaluation metric. The average evaluation results are provided in Table. \ref{table4} with red color indicating the best performance. Clearer illustrations of the evaluation results are shown in Fig. \ref{fig12}. According to the figure, the proposed 6DoF mesh saliency detection algorithm achieves the lowest KL score and SE score for most of the meshes when compared with the other methods. It also achieves the highest CC score for nearly half of the meshes under CC evaluation. The quantitative evaluation results indicate better performance of the proposed approach for 6DoF mesh saliency detection. It can also be observed from the figure that directly applying 3D saliency detection algorithms to 6DoF based tasks by considering the visibility of the points using 6DoF information might be impractical and the evaluation scores are relatively poor, indicating quite unsatisfying saliency detection results. These experimental results will be used as benchmarks for the presented 6DoF mesh saliency database.

\section{Conclusion and perspectives}
In this paper, we developed a setup for 3D eye-tracking experiment in 6DoF which allows subject's free movement during mesh observation. Based on this setup, we proposed a new 6DoF mesh saliency database which includes both the 6DoF data and eye-movement data from 21 subjects on 32 3D meshes. By analyzing the collected data, we obtained the statistical characteristics of the database and investigated the subject's visual attention bias during observation. We also observed the inter-observer variation which suggests that similar fixation patterns are more likely to appear for the same mesh than for different ones and investigated the viewing direction dependence which indicates a significant correlation between FDM similarity and the difference in viewing directions. Besides the analysis, we proposed a novel 6DoF mesh saliency detection algorithm using the uniqueness measure and  bias preference in 6DoF and designed an evaluation metric accordingly. The experimental results with 5 state-of-the-art 3D saliency detection methods demonstrate the superiority of the proposed approach and provide benchmarks for the presented 6DoF mesh saliency database.

The database will be made publicly available for research purpose to encourage more
work related to 6DoF based visual attention behaviour analysis and boost the development of mesh saliency detection in 6DoF.



\normalsize
\bibliography{main}


\end{document}